\documentclass[10pt,journal,compsoc]{IEEEtran}
\usepackage{booktabs} % For formal tables
\usepackage{multirow}
\usepackage{threeparttable}
\usepackage{ulem}

\usepackage[ruled,linesnumbered]{algorithm2e}% Metadata Information
\usepackage{multicol}
\usepackage{multirow}
\usepackage{graphicx}
\usepackage{subfigure}
\usepackage{xcolor}
\usepackage{colortbl,booktabs}
\usepackage{bbding}
\usepackage{booktabs}
\newcommand{\biaoti}{\fontsize{21pt}{\baselineskip}\selectfont}

\begin{document}

\title{\biaoti A Comprehensive Survey of Few-shot Learning: Evolution, Applications, Challenges, and Opportunities}

\author{
	Yisheng Song, Ting Wang$^\ast$, Subrota K Mondal, Jyoti Prakash Sahoo
	
	\thanks{(\textit{Corresponding author: Ting Wang.})}
\IEEEcompsocitemizethanks{
	\IEEEcompsocthanksitem Yisheng Song and Ting Wang are with the Engineering Research Center of Software/Hardware Co-design Technology and Application, Ministry of Education; the Shanghai Key Laboratory of Trustworthy Computing;  East China Normal University, Shanghai 200062, China (email: 71205902054@stu.ecnu.edu.cn; twang@sei.ecnu.edu.cn).
	% \IEEEcompsocthanksitem Wanli Chang is with the College of Computer Science and Electronic Engineering, Hunan University, Changsha 410082, China (email: wanli.chang.rts@gmail.com)
	\IEEEcompsocthanksitem Subrota K Mondal is with the Faculty of Information Technology, Macau University of Science and Technology, Macao, China (email: skmondal@must.edu.mo).
	\IEEEcompsocthanksitem Jyoti Prakash Sahoo is with the Department of Computer Science \& Information Technology, Institute of Technical Education and Research, Siksha ‘O’ Anusandhan University, India (email: jpsahoo@ieee.org).
}	}
%\author{Yisheng Song}
%\email{71205902054@stu.ecnu.edu.cn}
%%\orcid{1234-5678-9012}
%\author{Ting Wang}
%\authornote{Corresponding author: Ting Wang.}
%%\authornotemark[1]
%\email{twang@sei.ecnu.edu.cn}
%% \author{Mingsong Chen}
%% \email{mschen@sei.ecnu.edu.cn}
%\affiliation{%
%  \department{Shanghai Key Laboratory of Trustworthy Computing, Software Engineering Institute}
%  \institution{East China Normal University}
%%  \streetaddress{3663 N. Zhongshan Rd.}
%  \city{Shanghai}
%  \country{China}
%  \postcode{200062}
%}
%\author{Subrota K Mondal}
%\email{skmondal@must.edu.mo}
%\affiliation{%
%  \department{Faculty of Information Technology}  
%  \institution{Macau University of Science and Technology}
%  \city{Macao}
%  \country{China}
%%  \postcode{}
% }
% \author{Jyoti Prakash Sahoo}
%\email{jpsahoo@ieee.org}
%\affiliation{%
%  \department{Department of Computer Science \& Information Technology, Institute of Technical Education and Research}  \institution{Siksha ‘O’ Anusandhan University}
%  \city{Odisha}
%  \country{India}
%%  \postcode{}
% }

\IEEEtitleabstractindextext{%
\begin{abstract}

Few-shot learning (FSL) has emerged as an effective learning method and shows great potential. Despite the recent creative works in tackling FSL tasks, learning valid information rapidly from just a few or even zero samples still remains a serious challenge. In this context, we extensively investigated 200+ latest papers on FSL published in the past three years, aiming to present a timely and comprehensive overview of the most recent advances in FSL along with impartial comparisons of strengths and weaknesses of the existing works. For the sake of avoiding conceptual confusion, we first elaborate and compare a set of similar concepts including few-shot learning, transfer learning, and meta-learning. Furthermore, we propose a novel taxonomy to classify the existing work according to the level of abstraction of knowledge in accordance with the challenges of FSL. To enrich this survey, in each subsection we provide in-depth analysis and insightful discussion about recent advances on these topics. Moreover, taking computer vision as an example, we highlight the important application of FSL, covering various research hotspots. Finally, we conclude the survey with unique insights into the technology evolution trends together with potential future research opportunities in the hope of providing guidance to follow-up research.
\end{abstract}

\begin{IEEEkeywords}
	Cross-domain, Few-shot Learning, Fine Tuning, Meta learning, Transfer Learning.
\end{IEEEkeywords}

}
\maketitle

\section{Introduction}

%With the burgeoning development of 5G, artificial intelligence, big data, edge computing, cloud computing and other supporting technologies, the Internet of Everything (IoE) has evolved from a concept to a tangible and real-world implementation on the verge of changing the way we live. 
%An increasingly growing number of terminal devices are widely deployed in various critical infrastructures such as health, transportation, industrial production, environmental detection and home automation. The network transmits data without any form of interaction between humans and computers, humans and humans. It brings reliability and convenience to consumers, but it also opens up a new opportunity for intruders and introduces a unique and complex set of problems to the digital forensics field.

%\if Today is the age of the internet of everything. \fi
Recent advances in hardware and information technology have accelerated the interconnection of billions of devices in various IoT-enabled application domains. Smart and adaptive devices are increasingly deployed in critical infrastructures such as health, transportation, industrial production, environmental detection, home automation, and many other justifying the Internet of Everything (IoE) frameworks.
These massive number of terminal devices have been generating a huge amount of data, which need to be sent back to the server for central processing and storage. 
Although the total amount of generated data at the edge is very large, the volume of every dataset generated by a single device or single scene is extremely limited with very few samples.
%second for central processing, but the amount of data generated from a particular sensor is extremely limited. 
Traditional data-driven and single-domain algorithms do not perform well in these settings. To this end, numerous research has been conducted in exploring effective learning methods based on few samples and cross-domain scenes. Few-shot learning (FSL) as well as meta-learning have inevitably emerged as a promising way.
However, how to effectively obtain valid information from small sample data set or even cross-domain still remains the greatest challenge faced by FSL today.\if The network transmits data without any form of interaction between humans and computers, humans and humans. It brings reliability and convenience to consumers, but it also opens up a new opportunity for intruders and introduces a unique and complex set of problems to the digital forensics field. Although the network edge has a rich source of data, it still faces many challenges in practical applications. One of the biggest challenges is the heterogeneous and minimal data volume for individual terminals in edge scenarios, which is insufficient to support deep learning. \fi 

Besides, data distribution in real-world scenarios often has long-tail effects and it is difficult to generalize the same model across diverse domains. Taking the smart manufacturing industrial inspection as an example, such poor generalization issue has become one of the key challenges affecting the performance of its intelligent models.
%Primarily, this can be observed in the following areas. 
Specifically, current industrial quality inspection equipment requires certain specific lighting conditions, and the AI models trained under one lighting condition are difficult to "generalize" to other lighting conditions. In addition, considering the high accuracy requirements of industrial quality inspection scenarios, current AI models, usually using supervised learning, require a large number of defective samples for training. However, it is difficult to collect a sufficient number of satisfactory samples as the proportion of defective products in actual production scenes is small. Moreover, it is not possible to transfer across domains either. For example, a model for PC appearance defect detection cannot be directly used to detect defects of mobile phone screens, refrigerators, washing machines, or even different models of PCs. Similarly, as another example, when recognizing character images of components and circuit boards, as there are many suppliers of components, many types of devices, and many different character styles, thus it is not possible to collect a sufficient number of all kinds of samples of character images for one supplier, resulting in few or no samples of each type. 
Table \ref{tab:scences} provides a detailed summary of these challenges.

\begin{table*}
	\center
	\caption{Current challenges that industry urgently needs to address}
	\label{tab:scences}
	%\resizebox{\linewidth}{!}{
	\small
		\begin{tabular}{ccc}
			\toprule
			Scenes & Challenges & Key Solutions  \\
			\midrule
			Quality Inspection Line & \begin{tabular}[c]{@{}l@{}}Susceptible to light\\Few sample data\\Unable to transfer across-domains\end{tabular} & \begin{tabular}[c]{@{}l@{}}Few shot cross-domain transfer\\Robust model generalization\end{tabular} \\
			\hline
			Electronic Component Identification & \begin{tabular}[c]{@{}l@{}}A variety of colors, sizes and brands\\Insufficient samples of each brand\\Existence of unseen new brands\end{tabular} & \begin{tabular}[c]{@{}l@{}}Few-shot learning\\Transfer learning\\Unified feature representation\end{tabular} \\
			\bottomrule
		\end{tabular}
%	}
\end{table*}
%%% replaced "few-shot learning" with "FSL"
To address these challenges more effectively, FSL has produced some creative work on data, algorithms, and models. Up to now, as one of the most classical taxonomy, FSL is classified into meta-learning and metric-based learning. 
In this review, from the perspective of challenges, we divide the FSL into data augmentation, transfer learning, meta-learning, and multimodal learning. Data augmentation focuses on simulating data in different scenarios by metric or generative methods to maximize the actual data distribution. Transfer learning is mainly combined with pre-training and fine-tuning to extract prior knowledge from large-scale auxiliary data sets. When domain relevance is relatively uncommon or large auxiliary datasets are not available, transfer learning has definite limitations.  Meta-learning is currently the mainstream approach to solve the FSL problem. In recent years, some scholars have questioned "Is such a kind of meta or episodic-training paradigm really responsible and optimal for the FSL problem?". This has led to extensive discussions \cite{chen2019closer,doersch2020crosstransformers} on the necessity of meta-learning for FSL. As for multimodal learning, it integrates different dimensions of information, such as language, images, and audio. Multimodal learning is expected to break the dilemma of insufficient useful information for FSL in the real human information world.\if this is the first creative approach that has been used to address FSL by bridging the gap among different dimensional information to fundamentally enrich the core problem of few-shot data scarcity.\fi

%In this context, few-shot learning emerged, which was first proposed in \cite{miller2000learning} in 2000, and up to now, research on few-shot learning has undergone more than 20 years of evolution, with some creative work involving data, algorithms, and models. Few-shot learning in the traditional sense can be roughly divided into two types: meta-learning and metric-based learning. \textcolor{red}{ In this content, we further subdivide them into data augmentation, fine-tuning, meta-learning and multimodal fusion. Data augmentation is implemented directly through metric or generative methods to mimic different data, so as to reflect the real data distribution to the greatest extent. Metric-based learning and meta-learning have become popular directions in recent years, and many papers have shown that pre-training combined with fine-tuning can achieve competitive results. "Is such a kind of meta- or episodic-training paradigm really responsible and optimal for the FSL problem?" was presented, triggering the academic community \cite{bertinetto2018meta,chen2019closer,doersch2020crosstransformers} to rethink the necessity of meta-learning and episodic-training for few-shot learning. Meta-learning is mainly about achieving a good state of parameter initialisation by co-training on different tasks. Multimodal few-shot learning is one of the learning paradigms closest to human intelligence, and truly solves the few-shot learning dilemma by bridging the gap between different dimensions of information.}

Due to the specificity of FSL, each method of FSL is confronting multifaceted challenges to varying degrees. One of the most direct challenges in data augmentation is that the data samples are too limited and the model cannot evaluate the true data distribution by relying solely on a few samples. As a result, the model trained in this setting is biased and easily falls into over-fitting. In transfer learning, features can effectively alleviate the problem of FSL, where the volume of data is small and cannot be migrated across similar domains. Nevertheless, how to represent features effectively, how to reuse features between different tasks, and how to establish an effective mapping between data and labels are great challenges that exist in transfer learning. Moreover, in the meta-learning paradigm, when training the meta-learner with a set of tasks, it not only samples the data space but also the task space. By constantly adapting to each specific task, it makes the network have an abstract learning ability. When the training task and the target task are distinctly different, the effect of meta-learning is minimal. Furthermore, in the field of multimodal learning, 
extensive studies have been conducted to explore how to effectively integrate information from multiple modalities to assist the  FSL.
 
Several existing survey papers have investigated the related work of FSL, for instance, the work \cite{shu2018small} categorizes FSL approaches into experience learning and conception learning. The work  \cite{lu2020learning} classifies FSL approaches into generative models and discriminative models according to probability distribution. Lately, the work \cite{wang2020generalizing} proposes a new taxonomy to classify the FSL approaches from the aspect of data, models, and algorithms. Nevertheless, to the best of our knowledge, no one paper has ever provided a taxonomy from the perspective of challenges in FSL. 
%Such work is necessary and far-reaching. 
By summarizing the challenges of FSL, readers can better grasp the motivation and principle behind the FSL, rather than being limited to various models. A list of key acronyms used in this paper is summarized in Table \ref{Acronyms}.

\begin{table*}
	\centering
	\caption{A List Of Key Acronyms}
	\label{Acronyms}
%	\resizebox{\linewidth}{!}{
    \small
		\begin{tabular}{ll|ll}
			\toprule
            \multicolumn{4}{c}{NOMENCLATURE}        \\ 
			\midrule
			\textbf{Full Form} &  \textbf{Abbreviation} & \textbf{Full Form} &  \textbf{Abbreviation}\\
            \hline
			Artificial Intelligence & AI & Few-Shot Learning & FSL\\
			Deep Learning & DL & Machine Learning & ML \\
			Zero-Shot Learning & ZSL & One-Shot Learning & OSL  \\
			Neural Architecture Search & NAS & Conventional Neural Network & CNN \\
			K-NearestNeighbor & KNN & Support Vector Machine & SVM\\
			Nearestcentroid Classifier & NCC & Graph Few-Shot learning & GFL \\
			Variational Auto Encoders & VAE & Few-Shot Object Detection & FSOD \\
			Long Short-Term Memory & LSTM & Data Augmentation & DA \\
			Few-Shot Cross-Domain & FSCD & Contrast Learning & CL \\
			\bottomrule
		\end{tabular}
%	}
\end{table*}

\subsection{Organization of the Survey}
\begin{figure*}[h]
	\centering
	\includegraphics[width=\linewidth]{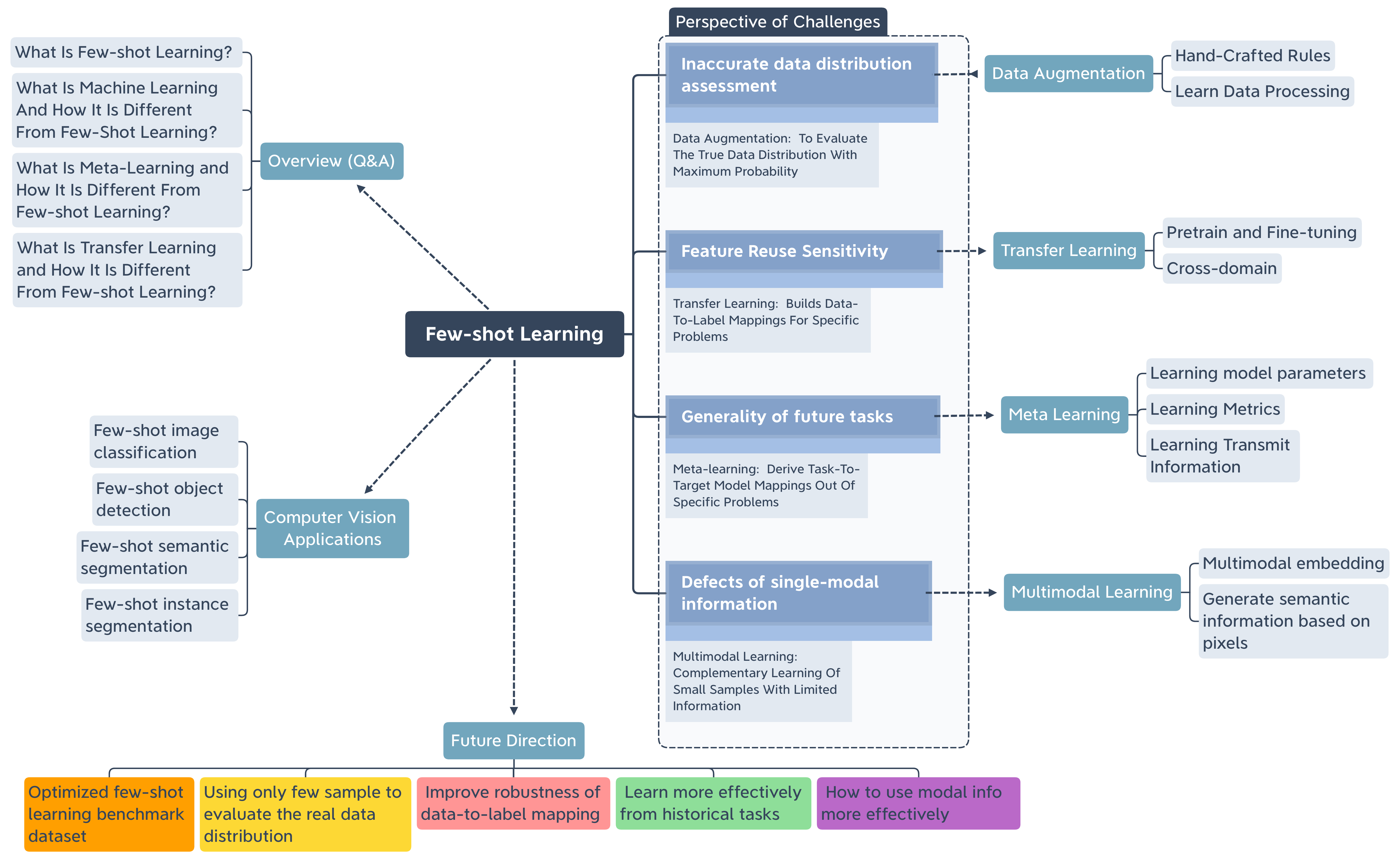}
	\caption{The conceptual map of the survey.}
	\label{fig:outline1}
\end{figure*}

The remainder of this survey is organized as follows.
Section \ref{secOverview} provides an overview of FSL, introducing FSL, comparatively analyzing machine learning, meta-learning, and transfer learning along with summarizing the current variants of FSL and challenges. 
Furthermore, to tackle the obstacles systematically, in this section we demonstrate a new taxonomy to classify the existing FSL related works. Section \ref{secDataAug} to Section \ref{secmul}
%, Section \ref{secTransferLearning}, Section \ref{secmeta} and Section \ref{secmul} 
present a systematic investigation on the current mainstream researches from the perspective of challenges in FSL and provide a comparative analysis from various aspects. 
%Depending on whether prior external knowledge is required, they are primarily divided into single-modal learning and multi-modal learning. Single-modal learning can be further divided into data augmentation, transfer learning, and meta-learning. They correspond to data, features and tasks, which means higher level of knowledge integration. 
With this taxonomy, a discussion and summary are provided at the end of each section, giving our insights into the respective fields accompanied by some potential research opportunities.
Section \ref{seccv} takes computer vision as an example, counting the latest progress of FSL in image classification, object detection, semantics segmentation, and instance segmentation in chronological order.  Section \ref{secop} delves into exploring the current challenges faced by FSL and how to seek breakthroughs in each branch. The overall outline of this paper is shown in Fig. \ref{fig:outline1}. 

The main contributions of this survey can be summarized as follows:
\begin{itemize}
	\item We start with the edge computing scenario, in which the few-shot learning challenges arise, explaining and clarifying several similar concepts that are easily confused with FSL. This will be beneficial to help readers establish the relationship between few-shot learning, transfer learning, and meta-learning.
	\item We comprehensively investigate the FSL related work from the perspective of challenges through knowledge graphs and heat maps. With this taxonomy, we divide the FSL into several different levels, where the highest level is multimodal learning that mainly uses various semantic knowledge to assist judgment, and the second, third, and fourth levels are single-modal learning that addresses data level, feature level, and task level challenges, respectively. Notably, we also provide insightful discussions on FSL cross-domain research, which is currently the more challenging direction in the field of FSL.
	\item We investigate an adequate number of papers in recent three years and summarize the main achievements of FSL in the field of computer vision, including image classification, object detection, semantic segmentation, and instance segmentation. 
	\item With these challenges mentioned at the end of the survey, combined with practical applications, we delve into the current challenges of FSL and explore how to find breakthrough points in each branch to jointly drive the research of FSL towards a more practical direction.
	\item We provide unique insights into the evolution of FSL and identify several future directions and potential research opportunities concerning each challenge.
\end{itemize}

\section{Concepts and Preliminaries}
\label{secOverview}
As a branch of machine learning (ML), FSL is still a young field. What is FSL, and how does it relate to machine learning, transfer learning, and meta-learning? What are the variants of FSL that currently exist? What benchmark datasets frequently appear in research papers? In this section, we will address the obstacles to FSL for readers by answering these questions.

\subsection{What Is Few-Shot Learning?}
The concept of FSL is inspired by the robust reasoning and analytical capabilities of humans, and it is widely found in edge computing scenarios. In 2020, Wang et al. \cite{wang2020generalizing} give a detailed definition of FSL through experience, task, and performance of machine learning, which is one of the most recognized definitions so far: A computer program is said to learn from experience
$E$ with respect to some classes of task $T$ and performance measure $P$ if its performance can improve with $E$ on $T$ measured by $P$. It is worth mentioning here that $E$ in FSL is very small. In recent years, relevant neural scientific evidence \cite{tulving1985many,tulving2002episodic} has shown that innate human abilities are related to various memory systems, including parametric slow learning neocortical systems and non-parametric fast hippocampal learning systems, which correspond to FSL's data-based slow learning and feature-based fast learning, respectively.

To better understand FSL, it is necessary to introduce two concepts, one is $N$-way-$K$-shot problem and the other is cross-domain FSL. The $N$-way-$K$-shot problem is often used to describe the specific problems encountered by FSL. In this case, the support set represents the small dataset used in the training phase, which generates reference information for the second phase of testing. The query set is the task on which the model actually needs to predict.
Notice that the query set classes never appear in the support set. Classical $N$-way-$K$-shot represents support set with $N$ categories and $K$ samples per category,  then the whole task has only $N$ $*$ $K$ samples. As thus, $N$-way-$1$-shot represents one-shot learning and $N$-way-$0$-shot represents zero-shot learning. The concept of cross-domain originates from transfer learning, which refers to the transfer the knowledge from source domain to target domain. There usually exist domain gaps between these domains. Cross-domain FSL integrates the features of cross-domain and FSL, and is a challenging direction that has recently emerged.

At this stage, there still exist many challenges in FSL, which are generated from various aspects, including but not limited to data, models, and algorithms. In this context, the challenges can be generally summarized according to the degree of integration of knowledge as follows:%The most direct challenge is that a few-shot sample cannot simulate the actual data distribution behind the limited data, thus falling into the few-shot data risk preference. On this basis, to a certain extent, feature reuse in high-dimensional space alleviates the problem of small sample data volume. Feature reuse can be relatively easy to transfer in the source and target domains. Still, it is only adequate for the current task and does not help for future missions.  Meta-learning explores the distribution of different tasks' distribution and uses the least training cost to obtain the fastest migration capability. When information in one dimension is missing, we can discuss using data from other measurements to compensate. Few-shot learning in multimodal fusion scenarios is expected to solve the ultimate challenge once and for all. Many papers have now made different levels of exploration for these challenges, and this lays down the four major approaches to address few-shot learning challenges in the following.
\begin{itemize}
	\item {\verb|Inaccurate data distribution assessment|}: FSL does not have access to large datasets due to costs, ethical, legal, or other reasons. Consequently, relying on few samples for learning produces biases in estimating the actual data distribution, which may be fatal for some tasks. To this end, maximizing the exploration of data distributions with limited information becomes the most significant challenge for FSL. Data augmentation is the direct way to address the inaccurate estimation of FSL. The primary efforts currently focus on exploring migratable intra-class or inter-class features and customizing specific images using generators.
	\item {\verb|Feature reuse sensitivity|}: Continuous accumulation of a priori knowledge by sampling large-scale auxiliary datasets. Transfer learning can easily use it from the source domain to the similar target domain. \if As key techniques of transfer learning, pre-training and fine-tuning are the primary methods used to learn the data-to-label mapping relationships.\fi Pre-training aims to extract high-dimensional feature vectors through a feature extractor, while the goal of fine-tuning is to make minor adjustments to the initial parameters of the pre-training. Transfer learning focuses on the data level and obtains more valuable features independent of the task by mapping data to labels. It has an outstanding performance in optimizing specific tasks, but it is generally limited by the characteristics of current tasks and has a poor generalization to future tasks. Especially when there is a large shift in the domain, without filtering and alignment of features may result in negative knowledge transfer.
	\item {\verb|Generality of future tasks|}: 
	Differing from transfer learning, meta-learning learns to quickly build mappings from known tasks to target models in previously unseen tasks by double sampling the task and data. In FSL, by exploring the task space, summarizing meta-knowledge in different tasks can result in fast aggregation of unseen tasks at a lower cost. As a general learning framework, meta-learning is independent of specific problems and more oriented to future tasks instead of optimizing the current one. However, meta-learning has proven effective only when the testing and training tasks are relatively similar, and it is highly depends on the network structure and lacks flexibility. When training meta-learners with a set of tasks at the same time, it is even difficult to adapt to the distribution of tasks, requiring a redesign of the network structure.
	\item {\verb|Defects of single-modal information|}: It is difficult to learn features effectively because FSL is inherently information-limited. This situation is improved to a great extent when aided by getting information from other modalities. In this respect, semantic assistance \cite{wang2020large,xing2019adaptive} is an excellent method to provide external prior knowledge, where through the introduction or generation of semantic information as weak supervision, adaptive classification can be accomplished in conjunction with the original model.
\end{itemize}

%As a branch of machine learning, few-shot learning is still in a relatively new stage of development. There are many variations of few-shot learning, including zero-shot learning \cite{wang2019survey}, one-shot learning \cite{targ2016resnet}, and short resource learning. In the field of machine vision, the main application of few-shot learning in the direction of computer vision is image classification, object detection, semantic segmentation, and instance segmentation. They have emerged many excellent works in the field of few-shot learning. The experimental analysis is presented in  Table \ref{tab:tasks} below. 

\subsection{What Is Machine Learning And How It Is Different From Few-shot Learning?}
%In essence, machine learning enables computers to simulate human learning behaviours, including vision, language, and reasoning capabilities. The machine attempts to learn and acquire knowledge and skills automatically by imitating human beings, continuously improve performance, and finally comprehend artificial intelligence. What is machine learning, why does it have such a magical power, and the difference and connection between machine learning and few-shot learning? These questions will be answered in this section.

The traditional Von Neumann computer architecture allows users to execute a series of instructions step by step in the form of a program \cite{mehrabi2021survey}. However, this method does not work in ML. On the contrary, ML uses large-scale datasets as input. Its judgment on a new sample is based on statistical results extracted by historical data. Now, the burgeoning of 5G \cite{painuly2021future} provides massive connectivity for millions of end devices, enabling an interconnection of everything.  
The total amount of data generated by terminal devices is huge, but the amount of one single data set is extremely small. Hence, traditional ML, whose performance strongly depends on large data sets, cannot perform well in this setting with few samples. To this end, FSL emerges and provides a promising way to handle the data scarcity scenario.

%Compared with the traditional centralized computing, the self-calculating terminal can perform real-time detection. However, with the development of machine learning to the present, its limitations have also been reflected. Due to the excessive dependence on data, it cannot handle few-shot samples under high concurrency.

\begin{figure}[h]
	\centering
	\includegraphics[width=\linewidth]{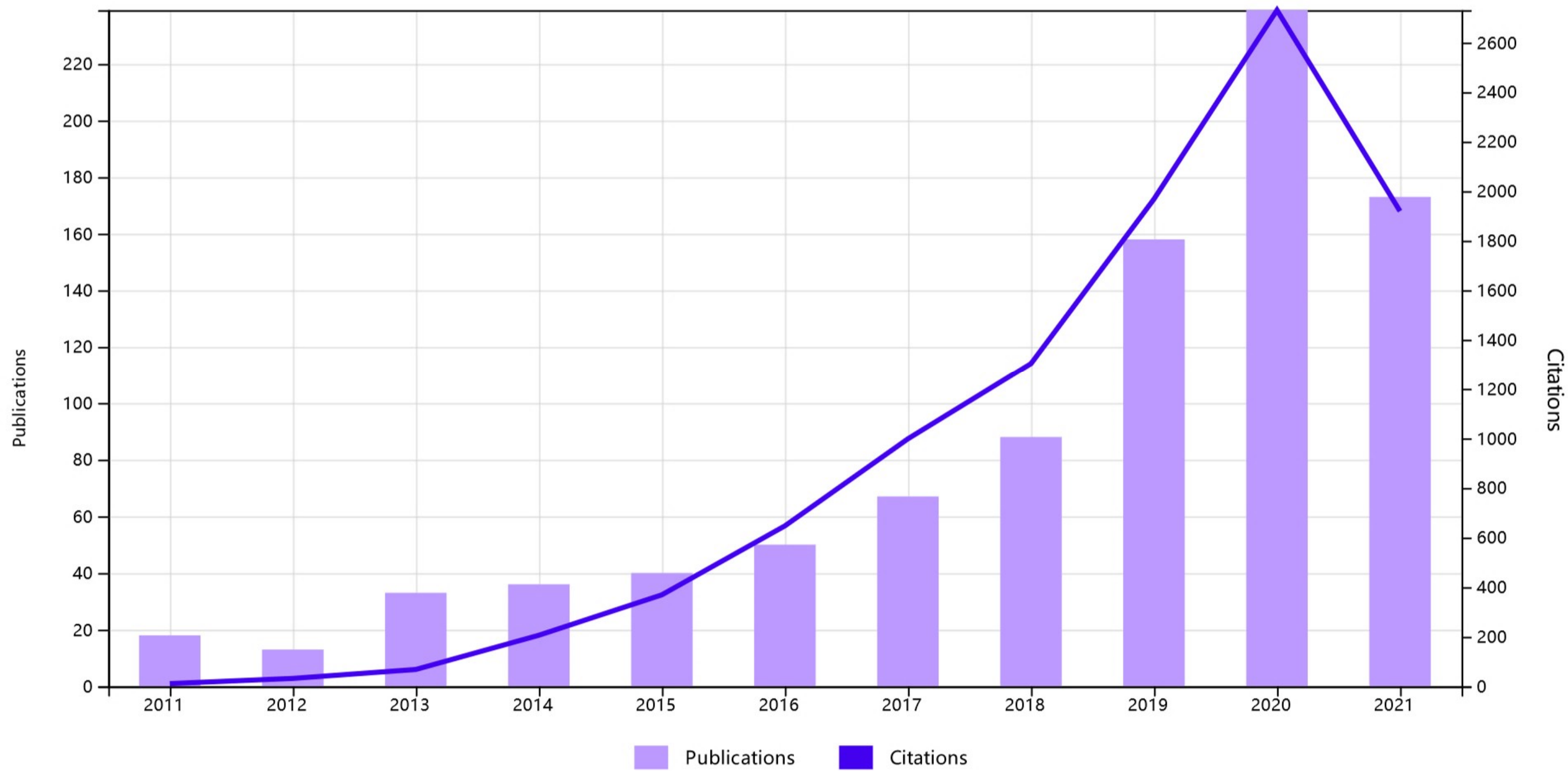}
	\caption{The number of FSL-related papers published in prestigious journals from 2010 to the first half of 2021, excluding citations.}
	\label{tab:papers}
\end{figure}

\begin{figure}[h]
	\centering
	\includegraphics[width=\linewidth]{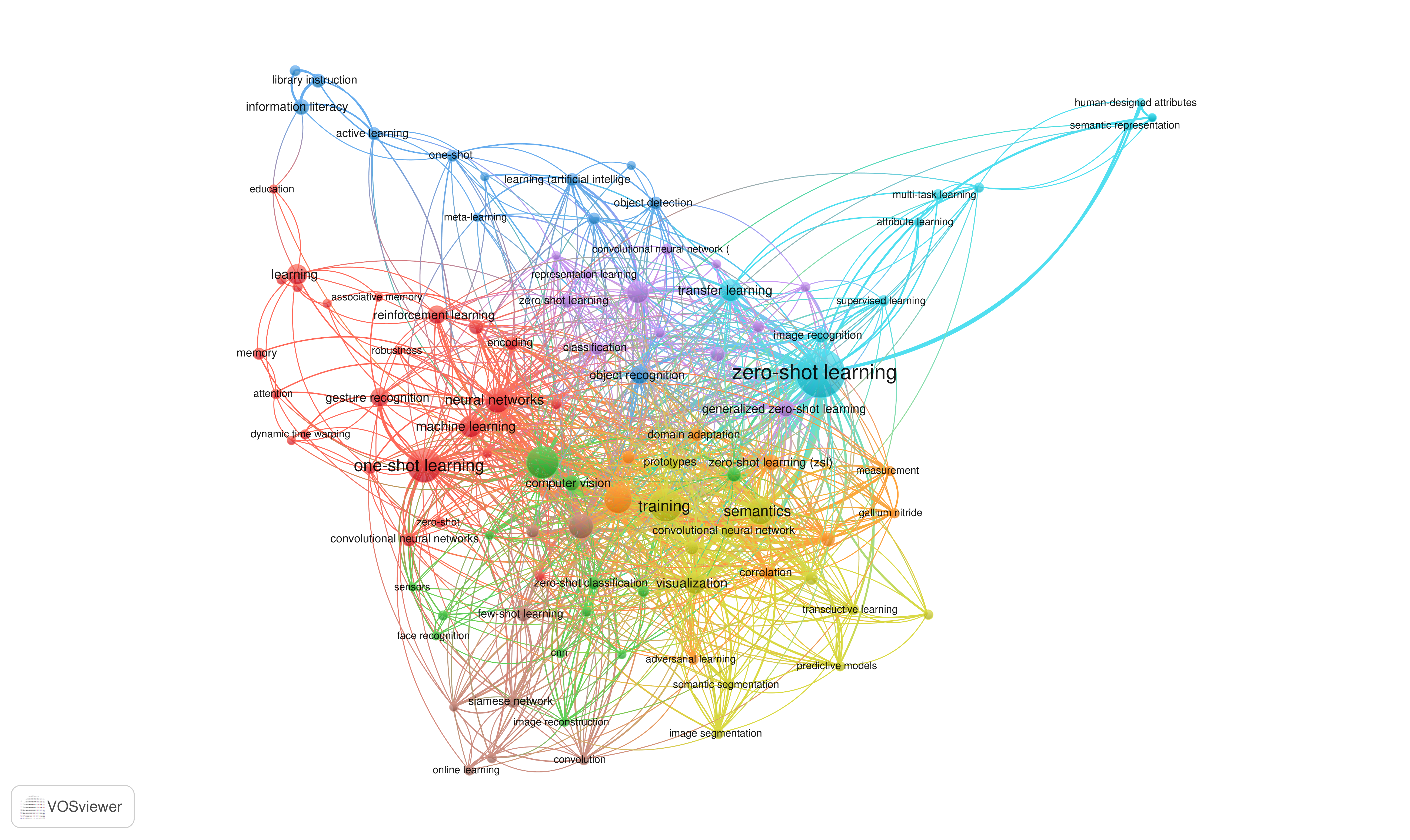}
	\caption{The knowledge graph uses Few-shot learning, One-shot learning and Zero-shot learning as keywords to relate the main advances and research directions in the field of Few-shot learning in recent three years. }\label{knowledge}
\end{figure}

In recent ten years, the research on FSL has been extensively conducted, and significant research progress has been achieved, e.g. the KGBert \cite{yao2019kg} proposed by Alibaba surpasses humans in the field of FSL for the first time. 
%We have investigated the number of core collection research papers in the past ten years with the themes of few-shot learning, one-shot learning, and zero-shot learning. 
Fig. \ref{tab:papers} exhibits the statistics of paper publications related to FSL in recent ten years based on the statistical results of the Web of Science.
As revealed, there are relatively few related papers from 2011 to 2015 due to that the FSL theory is still incomplete. With the rise of deep learning since 2015, the number of FSL related research papers has increased linearly almost every year. In the past 2020, the number of relevant papers has reached as high as 239, and the number of citations has reached 2731 times, accordingly.
%%Source of papers and number is not clear ;may require to mention sources (list of mainstream journal publishers).
Fig. \ref{knowledge} provides a knowledge map covering the hot research topics and cutting-edge developments in the field of FSL in recent years, including but not limited to zero-shot learning, one-shot learning, transfer learning, multi-task learning, and meta-learning. 
FSL related tasks include feature representation, visualization, robotics and cross-domain transfer. Among them, domain adaptation is a widely utilized method for few-shot cross-domain learning. Computer vision with predominant green color is the most active research field, including image classification, object detection, semantic segmentation, and instance segmentation.

The most significant difference between FSL and traditional machine learning is that the set of classes of the support set and the query set are disjoint. In machine learning, the classes of the test set are included in the training set in advance. FSL combines the limited supervision information with prior knowledge to train the model. The input of the model is generally given in the form of tasks. Through continuously collecting tasks, the model can recognize the similarities and differences between data as well as task. When the model encounters an unseen task, knowledge transfer can be accomplished quickly with just a few iterative training steps with appropriate initialization parameters. 
In contrast, traditional machine learning requires optimization through the loss function generated by a large scale data sets in the model. In conclusion, FSL is only a very young branch of machine learning, which mainly addresses the issue of difficult access to quality data sets in machine learning scenarios.

\subsection{What Is Transfer Learning and How It Is Different From Few-shot Learning?}
%There are many issues in the real world regarding datasets. Transfer learning explores how to make full use of previously labelled data for learning new tasks. Transfer learning aims to apply knowledge or patterns learned in a particular field or task to different but related fields or problems. There are two fundamental concepts in transfer learning, one is the domain, and another is the task. The domain can be understood as a specific scene at a particular moment. Tasks are what the model needs to do. Transfer learning can seek common knowledge between different fields and different tasks to reduce training costs.
Traditional transfer learning involves applying knowledge learned in the source domain to a different but related target domain.
In FSL, the limited amount of training data, domain variations, and task modifications are the key factors that cause the model to fail to transfer well from the source domain to the target domain. For instance, a medical image dataset with low similarity to the natural image dataset imagenet is difficult to identify accurately without the help of relevant expertise, even for a human-guided by only a few images. Certainly, it is also effective when the source domain and target domain are relative similar. The end in FSL tasks, if the prior knowledge is obtained from other tasks or domains by pre-training, FSL can belong to transfer learning, which mainly learns the mapping of data to labels.

According to the taxonomy of transfer learning, there are many variants of FSL problems, including one-shot learning (OSL), zero-shot learning (ZSL), and cross-domain few-shot learning.
\begin{itemize}
    \item {\verb|One-shot learning|}: OSL has only one correct label for each sample in the support data set, which aims to find the most similar class as a match among the seen-classes. During the police interrogation, these two processes are incredibly similar. The witness just looked once at the suspect, and the photos given by the police can be regarded as the query image. The witness only needs to answer 'yes' or 'no' towards those photos. Similarly, 
    one-shot learning does not classify the data specifically, but simply makes a cluster in order of similarity function. According to the existing work, one-shot learning can be divided into two main approaches. One is to use generative models to caste prior knowledge \cite{woodward2017active,mehrotra2017generative,chen2019image}, where bayesian programming learning \cite{salakhutdinov2012one} is the most representative framework \cite{schwartz2018repmet} in this field. Another method is to convert a OSL classification task into a verification task \cite{yoon2020oneshotda,jadon2021improving}. 
    
	\item {\verb|Zero-shot learning|}: ZSL was first proposed by Lampert et al. \cite{lampert2009learning}, which considers a more extreme case in FSL. In the absence of any query samples, the inference mechanism is solely relied on to identify samples that have not been seen before. ZSL is essentially done by using high-dimensional semantic features \cite{xian2018feature,kodirov2017semantic,naeem2021learning} to replace the low-dimensional raw data. Embedding representations and autoencoders are the most efficient ways to construct intermediate semantic spaces, which contains attributes that more comprehensively define the categories. Up to now, zero-shot learning is one of the closest methods to human intelligence that discerns previously unobserved categories. One-shot learning and FSL can essentially be considered as special ZSL. 
	
	\item {\verb|Cross-domian few-shot learning|}: In transfer learning, each class in the target domain has a sufficiently large number of available samples. When a large domain shift occurs between the source domains and target domains, knowledge transfer tends to become very challenging. Cross-domain few-shot learning combines the challenges of transfer learning and FSL. In the existence of domain gaps, where the intersection of classes in the source and target domains is empty, and the available sample size for each class in the target domain is extremely small. The improvement of the model's generalization capability through source domain data alone brings very limited performance to the model. The present work mainly focuses on the shift transformation of features and the construction of auxiliary datasets. Cross-domian few-shot learning can be regarded as one of the most challenging setting in the field of FSL at present.
\end{itemize}

\subsection{What Is Meta-Learning and How It Is Different From Few-shot Learning?}

\begin{figure}[h]
	\centering
	\includegraphics[width=\linewidth]{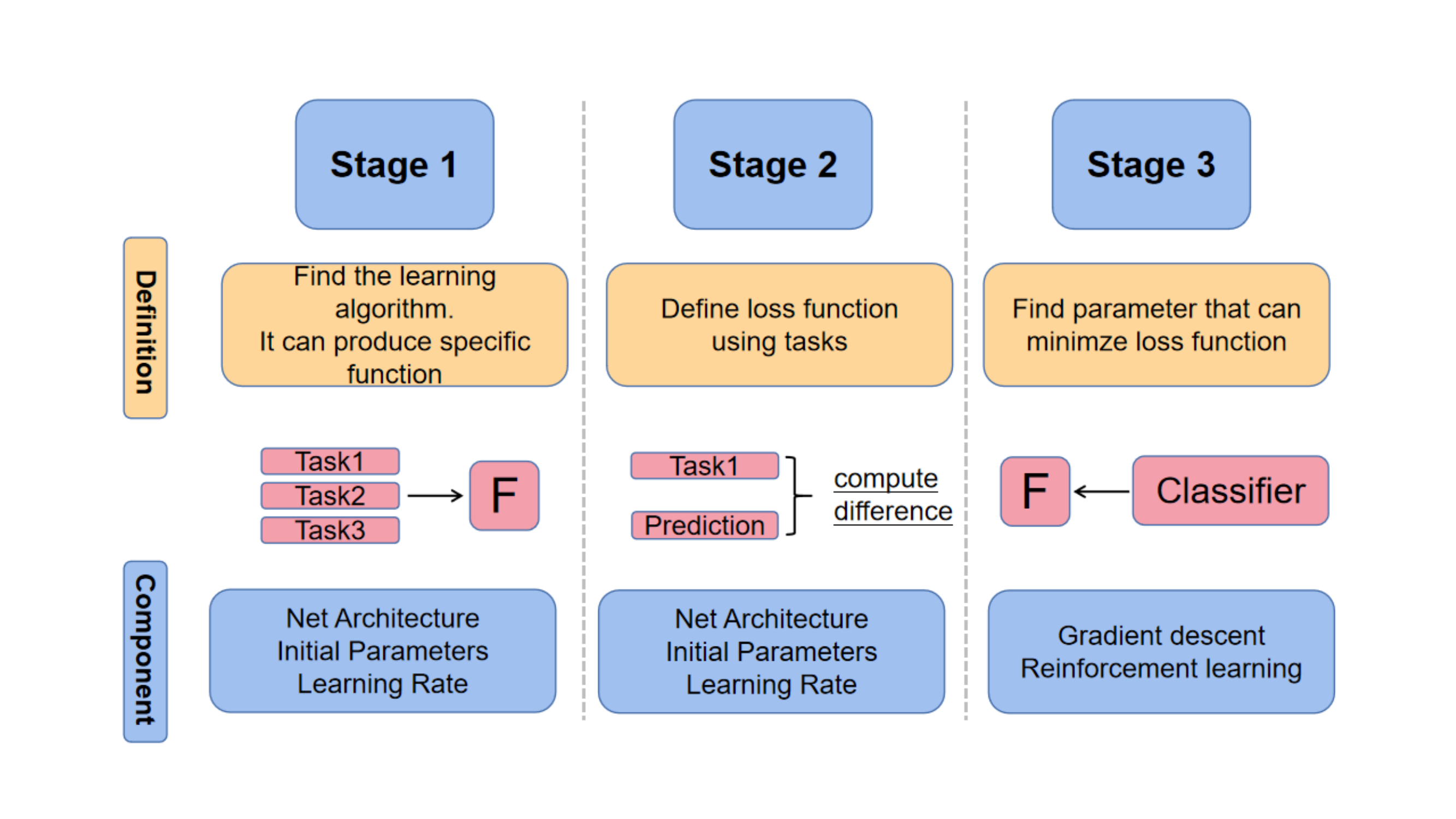}
	\caption{Meta-learning training three-step approach includes: find the learning algorithm, define loss function using tasks, find parameter that can minimze loss function. }
	\label{three}
\end{figure}

Meta-learning is a general learning paradigm that provides training on tasks in an episodic-training mechanism. Fig. \ref{three} illustrates the three-steps involved in meta-learning \cite{vanschoren2018meta} training. Meta-learning focuses on improving generalization for unseen tasks using prior knowledge. If prior knowledge is used to teach the model how to learn on a specific task, meta-learning can be regarded as a variant of FSL. It is emphasized that meta learning is not equivalent to FSL. FSL should be seen rather as an ultimate goal. It aims to achieve robust generations without relying on a large scale of datasets. By dual sampling of data and task space, meta-learning is enabled to construct a large number of auxiliary tasks related to the unseen task. Even if some papers do not use meta-learning, it is likely to improve the model's performance if episodic-training mechanism can be considered, such as meta reinforcement learning \cite{li2021mural,zhang2021meta}, meta video detection \cite{cheng2021meta}, and so on. 

Nevertheless, meta-learning has its own limitations: when the training and testing tasks exist obviously domain gap, Meta-learning is rarely used to initialize parameter weights. It can easily lead to negative migration of the model. In addition, meta-learning is highly dependent on the structure of the network, and needs to be redesigned for widely varying tasks. In spite of this, meta-learning is still one of the most effective methods to solve FSL.
%However, from the previous analysis, meta-learning is naturally suitable for few-shot learning, so a large number of few-shot learning papers are associated with meta-learning. Because meta-learning does not conflict with most other methods. 

\subsection{Datasets}
Before the availability of FSL benchmark datasets, researchers regularly used tasks like manually constructing N-way-K-shots to evaluate the performance of models. However, these simple tasks cannot reflect the complexity of real-world problems. After 10 years of evolution, the FSL benchmark dataset has completed the transition from a single domain, single dataset to a cross-domain, multiple dataset.

\begin{figure*}[h]
	\centering
	\includegraphics[width=0.85\linewidth]{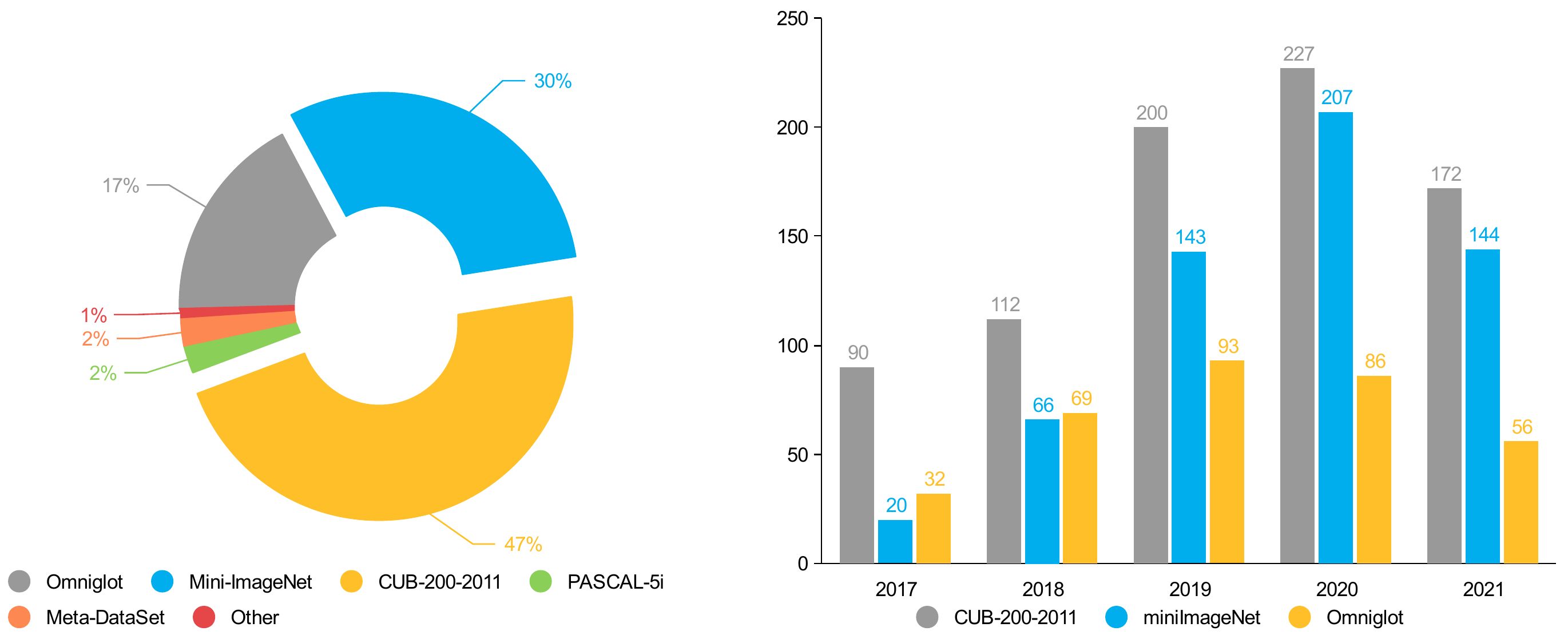}
	\caption{There are eight most frequently used datasets in few-shot learning, including the number of papers on mainstream benchmark datasets (2017-2021). There may be one paper that tests all mainstream benchmark datasets. Data from “paperswithcode” platform.} 
	\label{dataset}
\end{figure*}

\if This section investigated the FSL research papers collected by paperswithcode \footnote{Papers With Code: The latest in Machine Learning \url{https://paperswithcode.com/}} tool in the most recent five years. \fi 
As shown in Fig. \ref{dataset}, during 2017-2021, 898 papers used the CUB-200-2011 \cite{WahCUB_200_2011} dataset, accounting for 46.6\% of the total number of statistics; 587 papers used the Mini-ImageNet \cite{vinyals2016matching} dataset, accounting for 30.5\%; and 335 papers using the Omniglot \cite{lake2015human} dataset, accounting for 17.4\% ; 44 papers used the PASCAL-5i \cite{shaban2017one} dataset, and 46 papers used the Meta-DataSet \cite{triantafillou2019meta}. The other specific datasets are Paris-Lille-3D \cite{roynard2018paris}, N-Digit MNIST \cite{oh2018modeling}, SUN397 \cite{xiao2010sun}, which are used in 15 papers in the past five years. In terms of quantity, the CUB-200-2011, Mini-ImageNet, and Omniglot benchmark datasets occupy a dominant position in the field of FSL. Table. \ref{tab:command1s} compares the datasets mentioned above from different dimensions. By the publication of the article, a more objective benchmark dataset \cite{guo2020broader} for evaluating the cross-domain ability of FSL was proposed. 1) CropDiseases \cite{mohanty2016using}, a plant diseases
dataset , 2) EuroSAT \cite{helber2019eurosat}, a dataset for satellite images, 3) ISIC \cite{wang2017chestx}
a medical skin image dataset, 4) ChestX \cite{codella2019skin}, a dataset for X-ray
chest images. The similarity comparing to MiniImageNet is decreas across these datasets.
\begin{table*}
	\centering
	\caption{The latest performance of FSL in the main tasks of machine vision}
	\label{tab:command1s}
	\resizebox{\linewidth}{!}{
		\begin{tabular}{lllllll}
			\toprule
			Dataset Variant & Leader & Numbers/Classes & Train/Test & Content & Main FSL Task & License \\
			\midrule
			CUB-200 & Wah el al. \cite{WahCUB_200_2011} & 11788/200 & 5994/5794 & Birds & Few-shot Image classifition &  Attribution 4.0\\
			Mini-ImageNet & Vinyals et al. \cite{vinyals2016matching} & 600/100 & 480/120 & Real scene & Few-shot Image classifition& MIT \\
			Omniglot & Lake et al. \cite{lake2015human} & 32460/50 & 4800/1692 & Character & Few-shot Image classifition& MIT \\
			PASCAL-5i & Shaban et al. \cite{shaban2017one} & 576/20 & 11530/- & Really scence  & Few-shot Image classifition & MIT \\
			Meta-Dataset & Triantafillou et al. \cite{triantafillou2019meta} & 10 datasets & -/- & Really scence & Few-shot Image classifition & Multiple licenses \\
			BSCD-FSL & Guo et al. \cite{guo2020broader} & 4 datasets & -/- &
			Really scence, Satellite and medical image & Cross-domain few-shot learning & MIT \\
			Paris-Lille-3D& Roynard et al. \cite{roynard2018paris} & 450000000/50 & 450000000/380000000 & Point cloud  & Few-shot Semantic Segmentation & CC BY-NC-ND 3.0 \\
			N-Digit MNIST & Oh et al. \cite{oh2018modeling} & -/- & -/- & Character & Metric Learning & Apache License \\
			SUN397 & Oh et al. \cite{xiao2010sun} &  108,753/397 & 76128/21750 & Really scence & Few-shot Image classifition& - \\
			\bottomrule
		\end{tabular}
	}
\end{table*}

\subsection{Taxonomy}
\begin{figure}[h]
	\centering
	\includegraphics[width=0.95\linewidth]{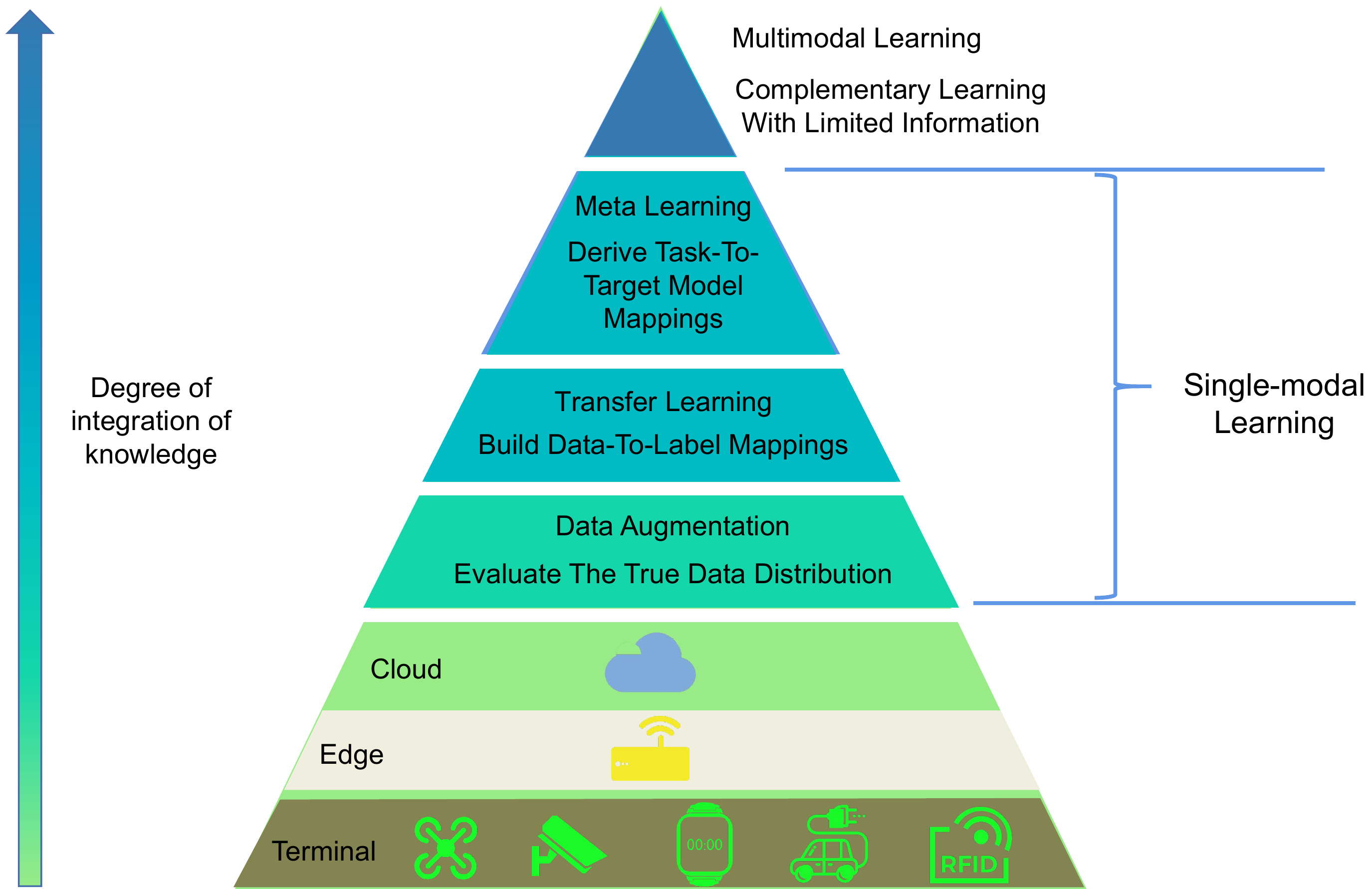}
	\caption{The entire taxonomy is presented in the form of a pyramid. The bottom level represents the "cloud-edge-terminal" edge computing scenario, which is characterized by few-shot real-time computation under high traffic. Based on this, the challenges of FSL are classified into four levels according to the degree of integration of the required knowledge. Among them, the challenges represented by data augmentation, transfer learning, and meta-learning are single-modal challenges. }
	\label{taxonomy}
\end{figure}

According to the degree of integration of knowledge, FSL is broaderly divided into a single-modal learning and a multi-modal learning. In this survey, The single-modal learning can be further divided into data augmentation, transfer learning, and meta learning. It mainly focuses on abstracting or transferring limited information into higher-level feature vectors or meta-knowledge. Multimodal learning is more close to the real world of human intelligence, which no longer relies on the limited sample, and tries to find the space of other modalities to assist the FSL. With this taxonomy, we exhaustively review and discuss each method. Fig. \ref{taxonomy} vividly demonstrates the FSL's taxonomy under the challenge perspective.

%Single-modal learning can be roughly divided into data augmentation, which evaluates the actual data distribution with maximum probability; transfer learning, which builds reusable data-to-label mappings for specific problems; and meta-learning, which derives task-to-target model mappings out of specific problems. Each of these three stages corresponds to the three main challenges of few-shot learning. %
%The few-shot learning taxonomy is explained in Fig. \ref{taxonomy} below.%

\begin{itemize}
	\item {\verb|Evaluate The True Data Distribution|}: The key to the difficulty of FSL is that limited samples cannot reflect the actual data distribution. The most intuitive idea of machine learning is to generate additional data based on a certain probability model or to extend the auxiliary data set using a large volume of unlabeled data from extending data. Existing work focuses on exploring feature differences that can be learned between classes or with external datasets at the semantic level. Handcrafted rules and automatic learning data processing are the two main approaches at this stage. 
	\item {\verb|Build Data-To-Label Mappings|}: Furthermore, if a large number of features from the benchmark dataset can be reused, which will significantly reduce the pressure of the model on the data. Pre-training and fine-tuning assist FSL by learning effective representation of data to labels, coupled with effective regularization of the underlying semantic features. In particular, the pre-training stage learns the optimal initialization parameters in a variety of different tasks, and the fine-tuning stage freezes most of the lower-level parameters and retrains only the parameters of the classification layer.
	\item {\verb|Derive Task-To-Target Model Mappings|}: Fine-tuning already has a good performance in baseline models with small samples. Nevertheless, in multi-task learning, a large scale of tasks are learned just as one task, which leads to a terrible generalization of the model. In contrast, meta-learning dual-samples the data and task space using episodic-training mechanism, finding latent associations between different tasks and thus having a good description of the whole task space. 
	\item {\verb|Complementary Learning With Limited Information|}:Multimodal learning has been proposed in deep learning for a long time, but it has only started to be integrated with FSL in recent years. Information in multimodal dimensions is closest to the real human information world, and it compensates to some extent for the inability of FSL to make accurate assessments of data distributions in a single modality. Pixels, semantics, and sounds can be supervised signals for FSL tasks, and even more recently unsupervised learning has been used to explore more robust feature representations using contrast learning.
\end{itemize}

\section{Data Augmentation To Evaluate The True Data Distribution With Maximum Probability}
\label{secDataAug}

In real-world FSL tasks, the number of samples in the support and query sets is usually limited due to privacy, collection costs, and labeling costs. To mitigate this issue, data augmentation is recognized as the most direct way to increase the sample richness in FSL. Nevertheless, the core risk of the FSL data augmentation is how likely the augmented dataset can evaluate the distribution behind the real data. Based on whether the data augmentation techniques can be reused on other tasks, FSL data augmentation is divided into hand-crafted rules and automatic learning data processing. %One of the main stages of hand-crafted rules includes data level and feature level. In Fig. \ref{data-level}, Fig. \ref{fea-level}, and Fig. \ref{process}, we show the data level, feature level, and learning data-process respectively.%

\subsection{Hand-Crafted Rules}
Hand-Crafted rules require guidance from experts with specialized domain knowledge. A representative result is that %X 
Bouthillier et al. proposed to randomly discard pixels \cite{bouthillier2015dropout} on a random rectangular area to generate black rectangular blocks of simulated noise. Similar operations also include random erase \cite{zhong2020random} and fill \cite{devries2017improved,yun2019cutmix} in FSL. Nevertheless, briefly relying on the simple transformation of single-sample pixels cannot prevent the risk of overfitting. Further, the hand-crafted rules contains data level and feature level according to the dimension of information. Table. \ref{tab:hand} summarizes the data augmentation methods for hand-crafted rules making.
\begin{table*}
	\caption{The latest performance of FSL in the hand-crafted rules.}
	\label{tab:hand}
	\resizebox{\linewidth}{!}{
		\begin{tabular}{llllccc}
			\toprule
			Model &  Core View & Key Approach & Experimental Dataset & Using External Dataset & Data Level & Feature Level \\
			\midrule
			FTT \cite{kwitt2016one} & Enriching Instant Attributes & Places-CNN & Transient Attributes Database & \CheckmarkBold & \CheckmarkBold & \\
			CSEI \cite{li2021learning} & Erase Repair & Metric based & miniImageNet & \XSolidBrush & \CheckmarkBold & \\
			Image Deformation \cite{chen2019image}  & Semantic Invariance & Meta-learning & miniImageNet & \CheckmarkBold & \CheckmarkBold &  \\
			AdarGCN \cite{zhang2021adargcn} & Denoising the collected web images & GCN layer & - & \CheckmarkBold & \CheckmarkBold  & - \\
			\begin{tabular}[c]{@{}l@{}}Covariance-Preserving Adversarial\\Augmentation Network \cite{gao2018low}\end{tabular} & \begin{tabular}[c]{@{}l@{}}"Variability" of covariance information\\as base instances\end{tabular} & Generative Adversarial Network  & ImageNet & \CheckmarkBold &  & \CheckmarkBold \\
			Spot and Learn \cite{chu2019spot} & extracts varying sequences of patches & reinforcement learning & miniImagenet & \XSolidBrush  &   & \CheckmarkBold \\
			Saliency-guided Hallucination \cite{zhang2019few} & Background - Prospective Learning & Realation network & miniImagenet & \XSolidBrush &  & \CheckmarkBold  \\
			Laso \cite{alfassy2019laso} & Explore the reliable differences between labels & Transfer learning & MS-COCO & \CheckmarkBold &   & \CheckmarkBold\\
			Dual TriNet \cite{chen2018semantic} & Semantic Synthesis Example & Auto Encoder & MS-COCO & \XSolidBrush &   & \CheckmarkBold\\
			\bottomrule
		\end{tabular}
	}
\end{table*}

\begin{figure}[h]
	\centering
	\includegraphics[width=\linewidth]{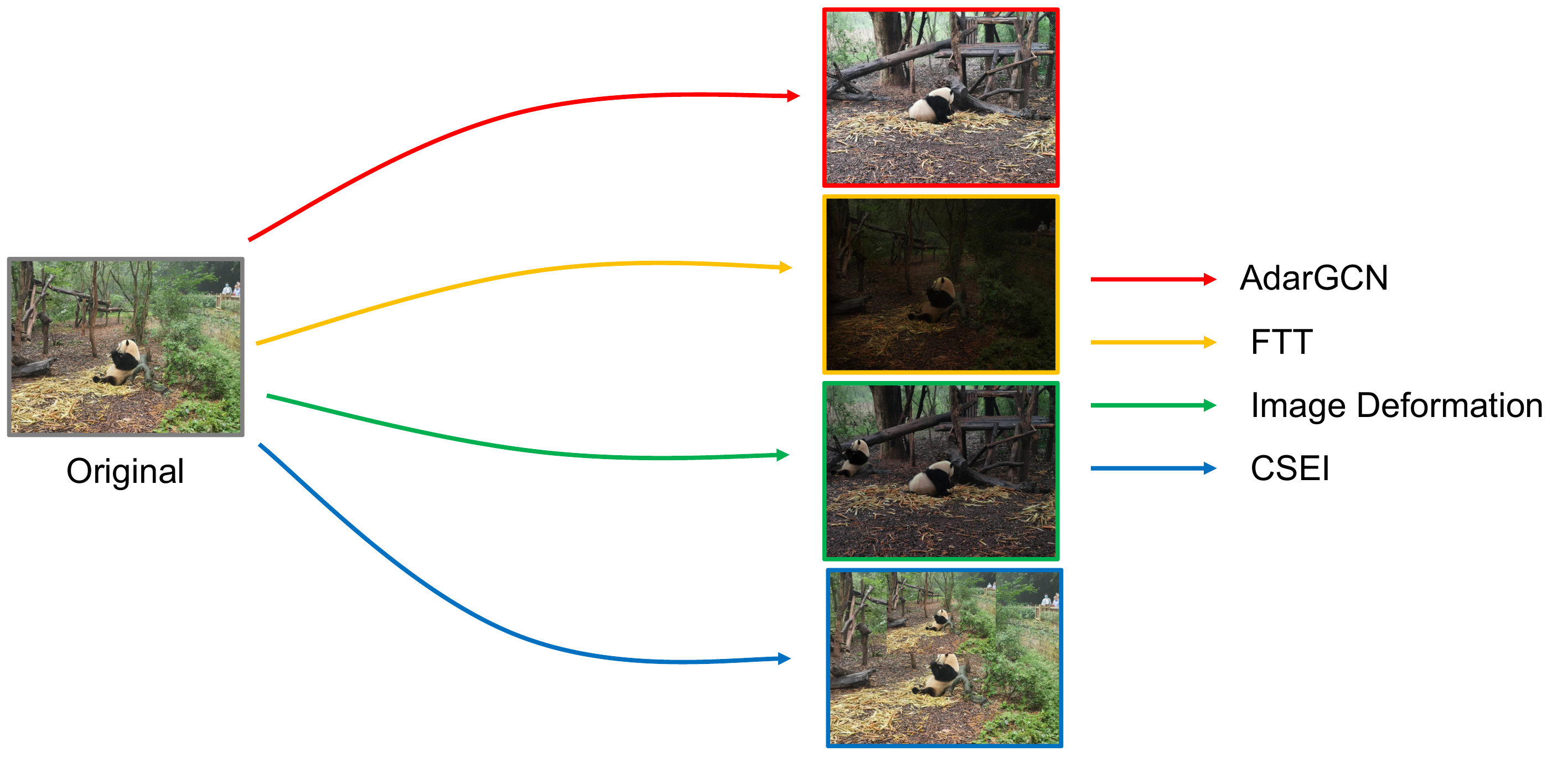}
	\caption{FSL data augmentation based on data level mainly includes internet data collection, environment variation, difference transfer and random crop filling. Here a picture of a panda is used as an example to implement the above variation. }
	\label{data-level}
\end{figure}

\subsubsection{Data Level}
The data level augmentation is mainly a transformation of the input data that aims to scale up existing data by making modifications to the data marginally for achieving diversity in model input. Random erasure \cite{zhong2020random} and random cropping \cite{devries2017improved,inoue2018data} are classical algorithms by simulating the images with different degrees of damage and thus improving the generalization of the model. Inspired by this, Li et al. \cite{li2021learning} discarded the traditional approach based on a entire feature extractor for images and instead focused on local patch images. These methods require large-scale datasets as support. It is not easy to achieve in the FSL settings. Conversely, CSEI \cite{li2021learning} does not require extra data sets. The specific operation is to erase most of the discriminatory area in the support set derived from the metric function and replace it with an image fill using the restoring operation. FTT \cite{kwitt2016one} enriches the data set by linear interpolation of some transiently transformed attributes, such as different weather and lighting. Z Chen et al. \cite{chen2019image} proposes an end-to-end approach to partitioning images as a whole inspired with the idea of MIXUP \cite{zhang2017mixup}, which argues that images preserve important semantic information even after they have undergone various distortions. The most significant difference between image distortion and GNN is that image distortion simply stitches two images together in a linear pattern. This method is able to achieve maximum deformation without loss of classification. In addition, it is a good direction for data expansion by using the large amount of unlabelled data sets in the real world for supplementation. Finally, when both the source and target classes both have only a limited number of samples, AdarGCN's \cite{zhang2021adargcn} implementation crawls data from internet resources and automatically removes irrelevant noise to achieve controllable data augmentation. At the same time, AdarGCN can automatically determine how far the information has been propagated in each graph node. In conclusion, data augmentation at the data level focuses on increasing the number of samples by means of pixel transformations and pixel generation. Fig. \ref{data-level} shows the main methods based on data level under hand-crafted rules.

\begin{figure}[h]
	\centering
	\includegraphics[width=\linewidth]{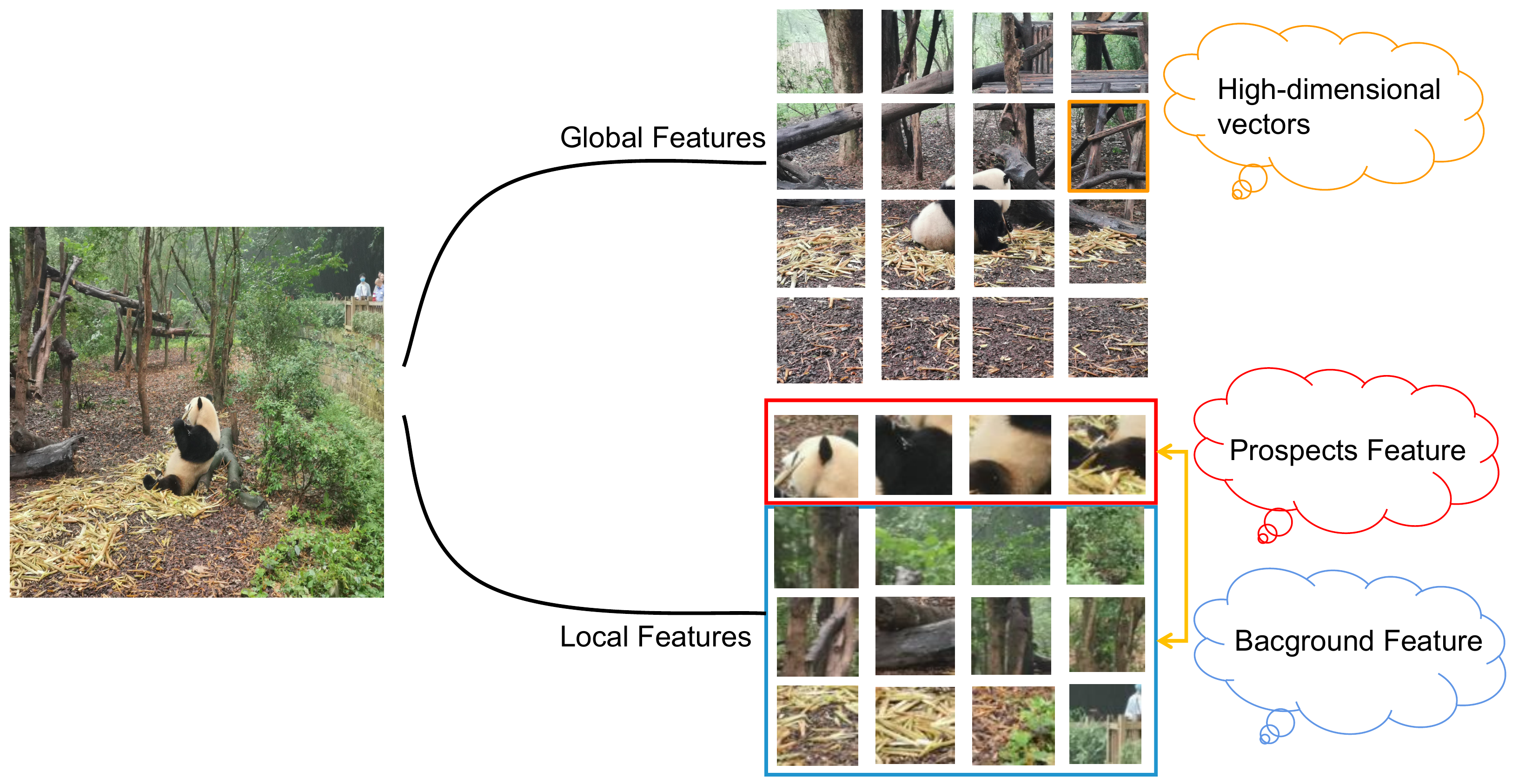}
	\caption{Feature level-based data enhancement can be mainly divided into global features and local features. Global features focus on the whole image, including the foreground and background. Local features, on the other hand, selectively focus on the subject part in the foreground.}
	\label{fea-level}
\end{figure}
\subsubsection{Feature Level}
Feature level data augmentation mostly maps pixel information into a high-dimensional latent space. It carries more valid information than mere original pixels. Gao et al. \cite{gao2018low} first explored the underlying distribution behind few-shot data and proposed an adversarial covariance augmentation network to overcome the limitations of FSL. Its experiments have shown that relying solely on learning the features of the entire image brings noise into the results. Chu et al. \cite{chu2019spot} tried to compute feature representations for each patch, rather than the entire image. Each small patch is connected by RNN and the features of the image are further fused. \if The features are connected in temporal order using an RNN and autonomously decided by reinforcement learning which uses patches of image for feature aggregation. \fi This heuristic algorithm is far superior to simple attentional models\cite{chen2003visual}.
Zhang et al. \cite{zhang2019few} explain partial feature learning from another perspective, proposing to use a pre-trained model to decompose visual features into three parts and then select the original, foreground, and background images to be re-stitched into new visual features. Similarly, Laso \cite{alfassy2019laso} explores the differences in features between different datasets in a high-dimensional space. Combining different labels through the intersection and complementation of sets allows images to contain key information from multiple classes at the feature level simultaneously. Training this part of the image as a support set can significantly improve the classification performance of small samples. Chen et al. \cite{chen2019multi} go further by extending the features to a high-dimensional semantic space. In FSL, feature-level augmentation is more effective than data-level augmentation by modeling the valid information in a compressed manner. Fig. \ref{fea-level} shows the main methods based on features level under hand-crafted rules.

\begin{figure}[h]
	\centering
	\includegraphics[width=\linewidth]{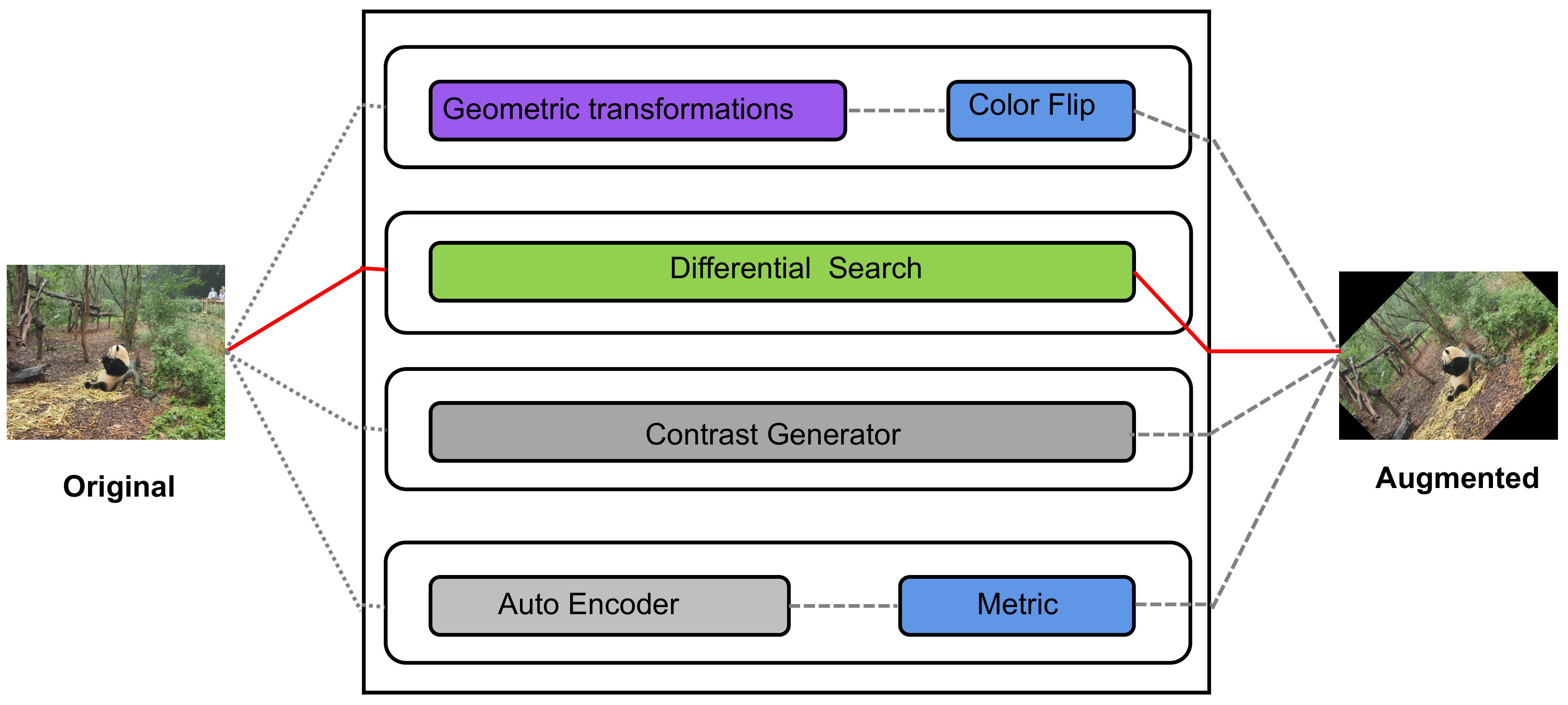}
	\caption{Learned Data Processing aims to learn a policy generator in multiple task spaces so as to automatically match different tasks. Its biggest benefit over hand-crafted rule is that it can be reused. }
	\label{process}
\end{figure}
\subsection{Learning Data Processing}
In 2018, data augmentation entered the area of auto augmentation with the maturation of meta-learning. Through the combination of meta-learning with other data augmentation methods, a large amount of excellent work emerged during this period. Hu et al. \cite{li2020dada} was inspired by the DARTS algorithm to abstract the data augmentation into multiple sub-strategies, each with a certain probability of being selected according to the different few-shot tasks. In addition to the probability-based method, another approach is based on generation. Li et al. \cite{kang2020cwgan} proposed the adversarial feature phantom network-AFHN. The phantom diversity and discriminative features are conditional on a small number of labelled samples. Chen et al. \cite{chen2019multi} attempt to train a meta-learner and generate a network to learn similarities and differences between images end-to-end by fusing pairs of images. MetaGAN \cite{ma2020metacgan}, on top of which an adversarial generator conditional on the task is introduced, which helps FSL tasks form generalizable decision boundaries between different classes. On the other hand, Zhang et al. \cite{zhang2019few} further demonstrate the usefulness of phantom data generation for FSL and propose a low-cost automated data generation method that uses a direct foreground-background combination to generate feature space-level data for training. In addition, it is also effective to explore the migratable differences between and within classes in support datasets. Delta-encoder \cite{schwartz2018delta} uses auto-encoder \cite{liou2014autoencoder} to learn differences in the same class for transfer learning, which is different from metric-based computation \cite{kim2019variational} of visual similarity. Table. \ref{tab:learn} summarizes the data augmentation methods for learning data process. Fig. \ref{process} shows the main approaches involved in auto-learning data processing under FSL.

\begin{table*}
	\centering
	\caption{The latest performance of FSL in the field of learn data process.}
	\label{tab:learn}
	\resizebox{\linewidth}{!}{
		\begin{tabular}{llllc}
			\toprule
			Model &  Core View & Key Approach & Experimental Dataset & Using External Data \\
			\midrule
			DADA \cite{li2020dada} & \begin{tabular}[c]{@{}l@{}}Automatic generation of \\ enhancement policies\end{tabular} & Gradient Descent  & CIFAR-10 & \CheckmarkBold \\
			AFHN \cite{kang2020cwgan} & Condition-based generation & Generative Adversarial Network  & MNIST & \CheckmarkBold \\
			MetaGAN \cite{ma2020metacgan} & Generate extral data & Generative Adversarial Network  & MNIST & \CheckmarkBold \\
			Delta-encoder \cite{schwartz2018delta} & Differential transfer &  Auto Encoder & Mini-ImageNet & \CheckmarkBold \\
			MSFN \cite{kim2019variational} & \begin{tabular}[c]{@{}l@{}}Calculate multi-scale features \\ and the similarity of \\each class of labels  \end{tabular} & Metric Learning & Omniglot & \CheckmarkBold \\
			\bottomrule
		\end{tabular}
	}
\end{table*}

\subsection{ Discussion and Summary}
To maximize the evaluation of the distribution of the real data in FSL setting, data augmentation has from the hand-crafted rules to the auto learned data processing stage. The watershed is the maturity of meta-learning in 2018. This section comprehensively investigates the emerging representative technologies in the field of data augmentation and reviews the evolution of few-shot data augmentation. Table. \ref{tab:learn} summarizes the model in different dimensions clearly.

%In the future, few-shot learning data arguments will inevitably develop in a more universal and inclusive direction. The argument strategy that only applies to specific datasets is no longer suitable for few-shot learning. Especially in zero-shot learning, the automatic learning and migration of algorithms under meta-learning will become the mainstream.

\section{Transfer Learning Builds Data-To-Label Mappings For Specific Problems}
\label{secTransferLearning}

\begin{figure}[h]
	\centering
	\includegraphics[width=0.9\linewidth]{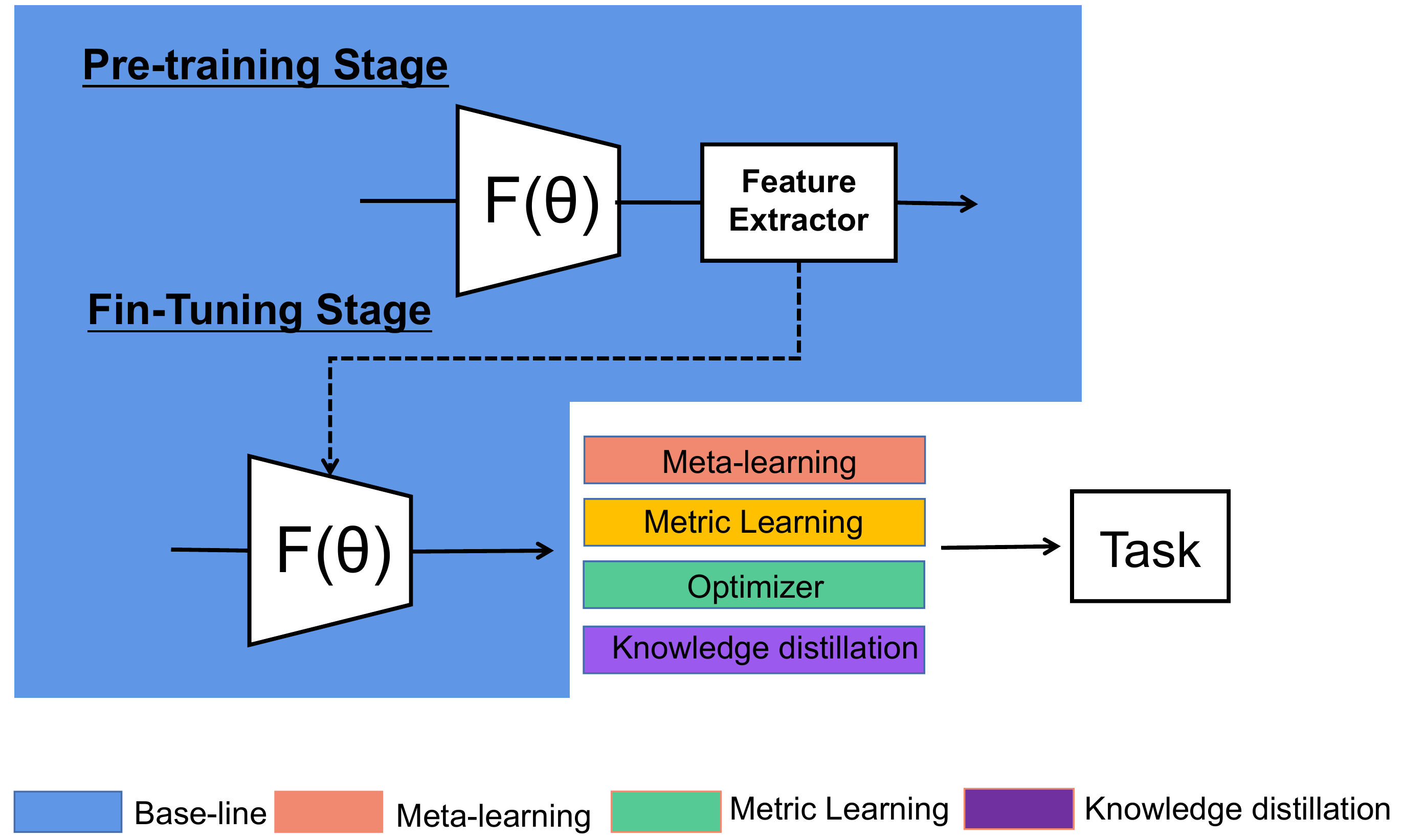}
	\caption{Transfer learning can be divided into pre-training and fine-tuning stages, where the baseline model can be combined with other techniques to improve the model performance. }
	\label{transfer}
\end{figure}

Transfer learning \cite{zhuang2020comprehensive} is a classical learning paradigm, which aims to solve the challenging problem that there are only a few or even no labelled samples \cite{kevin2021federated} in the FSL \cite{cai2020transfer}. Feature reuse is the core idea of transfer learning to solve FSL absence of data setting. The basic operation is to pre-train the model on an extensive dataset and then fine-tune on the limited support set. When source and target domains exist a large gap, knowledge transfer is invariably much less effective. This cross-domain setting brings a new challenge for FSL. In FSL, transfer learning can be broadly divided into  pre-training and  fine-tuning stage, which can also be referred to the baseline. Fig. \ref{transfer} illustrates the general process.

\subsection{Pre-training and Fine-Tuning}
From 2012 to 2018, a large number of excellent works have emerged in the field of computer vision and natural language processing, such as MobileNet \cite{howard2017mobilenets}, ResNet \cite{he2016deep}, ELMO \cite{peters2018deep}, GPT \cite{radford2018improving}, and BERT \cite{devlin2018bert}. In particular, the area of natural language processing was slow to progress before the advent of pre-training models. It has grown considerably under the leadership of BERT as computing power has increased and excellent pre-training models have been proposed. 

As a downstream task, how to use these excellent models to obtain features will largely alleviate the pressure on the data for FSL. Especially for few-shot image classification, as a pre-training model \cite{dhillon2019baseline,chen2020new}, it needs to use an external large-scale label dataset to extract prior knowledge from similar task. The most common practice is to design a backbone model without classifier layer, which includes convolutional neural networks or auto-encoders. The input of the model is an array of images, and the output is the feature vector embedding in a high-dimensional space \cite{hussain2018study}. High-dimensional feature vectors obtain sufficient valid semantic information about the target image. After the pre-training was put forward, the researchers proposed fine-tuning later. Most parameters in the pre-training are frozen, and only the classification layer parameters are updated in the testing stage. Many recent works \cite{nakamura2019revisiting,chen2019closer} have proved that fine-tuning can improve the 5-way-1-shot tasks accuracy rate by 2\%-7\% compared with the baseline model. Although the number of samples in the support set and query set is small, pre-training and fine-tuning are still very helpful for improving the accuracy of FSL. The conclusions are analogous in natural language processing as well. The authors in \cite{fabbri2020improving,lifchitz2019dense,yu2020transmatch} have also shown that fine-tuning can be embedded in state-of-the-art meta-learning or semi-supervised learning frameworks for optimizing model parameters.

Dhillon et al. \cite{lifchitz2019dense} replaced the standard activation function with cosine similarity, and Nakamura et al. \cite{nakamura2019revisiting} replaced conventional gradient descent with an adaptive gradient optimizer, which both improve the fine-tuning process in the accuracy of the model. Currently, fine-tuning is usually combined with meta-learning. Cai et al. \cite{cai2020cross} attempted to integrate them to train networks with specific layers. However, experimental results suggest that since the support and query sets do not overlap in FSL setting, transferring whole knowledge from the source dataset is not the best solution for FSL. Shen et al. \cite{shen2021partial} suggest that knowledge should be transferred specifically for parts. The degree of transferability needs to be controlled by freezing or fine-tuning specific layers in the backbone model. Similarly, fine-tuning can also be used to prevent new classes of networks from polluting the feature space of the basic classes. Up to now, the FSL and fine-tuning have been widely used in tasks like plant disease and insect pest identification \cite{argueso2020few}, road detection \cite{majee2021few}, and automatic question and answer \cite{ram2021few}.

\subsection{Cross-Domain Few-shot Learning}
The latest progress of FSL largely depends on the labelled data of the training stage. However, it is unrealistic to collect various forms of datasets for specific tasks in many practical applications, which results in challenging of FSL between intensely different domains. Cross-domain few shot learning integrates FSL and domain adaptive problems, which is a relatively comprehensive and challenging setting. For a long time, the benchmark datasets commonly used for FSL have suffered from a standardized dataset structure and large similarity of natural scenes, which leads to that models perform well on standard datasets but get unacceptable results in the real world task. Google first released a FSL cross-domain dataset named Meta-Dataset \cite{triantafillou2019meta} in 2020, which includes a total of 10 public image datasets including ImageNet, CUB-200-2011, etc. Yet these datasets are still focused on natural scenarios and cannot be broadly regarded as cross-domain few-shot benchmark datasets. Until the availability of BSCD-FSL \cite{guo2020broader} datasets. According to the degree of similarity with the ImageNet, it is divided into CropDiseases \cite{mohanty2016using}, EuroSAT \cite{helber2019eurosat}, ISIC \cite{wang2017chestx}, ChestX \cite{codella2019skin}. The authors extensively evaluate the performance of current FSL methods, and experiments show that the accuracy of all methods is correlated with the proposed natural image data similarity metric. Nowadays, cross-domain FSL focuses on distinguishing domain-irrelevant features and domain adaptive techniques with transfer learning.   
%%%%%%%\if mainly \fi

The objective of domain adaptation is to transfer knowledge from the source domain to the target domain, which has the same set of classes but a different data distribution than the source domain. Recently, much work has used adaptive networks to align their features with a new domain or to select domain-irrelevant features from multiple backbone model. Dvornik et al. \cite{dvornik2020selecting} obtained multiple domain representations separately by training a set of feature extractors with different domains. Setting the model to a dataset with multiple domains during training allows an attempt to migrate to other domains during the testing stage. Nevertheless, this approach may not be effective during the meta-training and meta-testing phases when the domains are orthogonal. Based on this, FRN \cite{wertheimer2021few} explores the potential space for few-shot image classification, using ridge regression to reconstruct and normalize the feature map without adding new learning parameters. FWT \cite{tseng2020cross} utilizes only the source data for the affine transformation of features, as do LRP-GNN \cite{sun2021explanation} and SBMTL \cite{rusu2018meta}. FD-MIXUP \cite{fu2021meta} constructs auxiliary datasets by mixup and uses encoders to learn domain-irrelevant features to guide the network generalization to other tasks. STARTUP \cite{phoo2020self} takes advantage of not only the source data but also assumes that the model has access to a lot of unlabeled target data during training. A large amount of unlabeled data is used to enhance the generalizability of the model to other domains. Metric-based approaches are frequently used for semi-supervised and unsupervised cross-domain FSL. A recent paper by Lu et al. \cite{liu2020universal} uses attention as a metric strategy to reweight and combine domain-specific representations. Chen et al. \cite{chen2020new} based on a meta-baseline by pre-training the classifier on all base classes and classifying a small number of samples based on the nearest centroid algorithms for meta-learning, which greatly surpasses the latest state-of-the-art methods. Li et al. \cite{li2021universal} inspired by \cite{bilen2017universal, rebuffi2017learning}, proposes to map domain-specific features to the same shared space, thus achieving a domain-irrelevant universal representation. 
%At present, fine-tuning techniques perform better than any other baseline for this problem of cross-domain learning. Chowdhury et al. \cite{chowdhury2021few} simply combining pre-training with L2 regularization also has achieved good performance. Riantafillou et al. \cite{triantafillou2021learning} propose a \textcolor{blue}{data-augmented} template that requires only inferring a small number of parameters to be inserted into the template to achieve parameter initialization for different tasks. SUPMOCO \cite{chen2021momentum} combines comparative and supervised learning, for instance, discrimination in a single framework, largely avoiding task-related irrelevant information is discarded. Choi et al. \cite{choi2019few}, inspired by multitasking learning, build multiple pools of embedded subnetworks to achieve task diversity without losing domain-invariant properties by sharing parameters of the underlying network.  

\subsection{Discussion and Summary}
While meta-learning methods have higher performance than transfer learning in standard FSL settings, the situation is reversed in cross-domain FSL settings. A newly published paper recently pointed out that the improvements from fine-tuning and pre-training are similarly very limited when the domains appear orthogonal. In the pre-trained feature space, the base classes form compact clusters, while the new classes are distributed in large difference groups. Currently, the actual deployment of trained models into production environments is often not adapted to rapidly changing environments. Pre-training can be seen as a task with many learning classes, but it is only a single learning task.

\section{Meta-learning Derive Task-To-Target Model Mappings Independent Of Specific Problems}
\label{secmeta}
Meta-learning learns historical prior knowledge from a dual sampling of data and tasks, then extracts meta-knowledge to apply to future tasks. Meta-learning is independent of the specific problem, and exploring an optimal initialization parameter in task space, discarding the task-independent feature representation under traditional supervised learning. Up to now, most of the meta-learning models are updated with parameters using traditional gradient descent. Absolutely, there are also non-gradient descent methods based on reinforcement learning and metric methods. In FSL, meta-learning can be used to automate the learning of model parameters, metrics function, and the transfer of information
\subsection{Learning Model Parameters}
Most of the deep learning frameworks use different parameter initialization methods, such as uniform distribution, normal distribution, and so on. The biggest problem with this random initialization is that it easily falls into the local optimal position. The goal of meta-learning is to train a hyperparameter generator, the classical methods being MAML \cite{finn2017model}, Repital \cite{nichol2018first} even their derived variants. MAML identifies the global optimization direction by calculating the optimization direction for each task. Compared to MAML, Reptile can update fewer parameters at once. The biggest difference between meta learning and multi-task learning is that multi-task learning only focuses on the performance of the current task. Meta-learning was demonstrated to perform better than transfer learning with a standard FSL benchmark dataset. Nevertheless, meta-learning is more sensitive to network structure and requires fine tuning of hyperparameters. After that more versions have evolved to address these issues separately. Such as MAML++ \cite{raghu2019rapid}, First-order MAML (FOMAML) \cite{nichol2018first}, Meta-SGD \cite{li2017meta}, TAML \cite{jamal2019task}, iMAML \cite{rajeswaran2019meta}, iTMAML \cite{rajasegaran2020itaml}. Of which Meta-SGD, expect MAML, finds the optimal learning rate and update the direction of the parameters at the same time, in addition to learning the initialization parameters. TAML \cite{jamal2019task} is a task-independent method, which overcomes the problem that MAML can only use an external model. Subsequently, IMAML \cite{rajeswaran2019meta} proposes a new loss function and a corresponding method for computing the gradient, making it possible to obtain the gradient of the parameters by calculating only the solution of the loss function, without caring its specific optimization method. iTMAML \cite{rajasegaran2020itaml} based on TAML, which implements automatic task recognition. It can be quickly adapted to new tasks by updating when the data is in a continuous state. At present, MAML has been widely used in various tasks \cite{joseph2021reproducibility,naman2021fixed,jeong2020ood}, producing different variants. Table. \ref{tab:learn2} distinguishes between MAML, Reptile and their variants in various perspective.

\begin{table*}
	\centering
	\caption{Summary MAML, Reptile and their variants}
	\label{tab:learn2}
	\resizebox{\linewidth}{!}{
		\begin{tabular}{lllcc}
			\toprule
			Model &  Directions for improvement & Key Approach & First order gradient & Two-step gradient \\
			\midrule
			MAML \cite{finn2017model}& Original & Inner-loop+outer-loop  & \XSolidBrush & \CheckmarkBold \\
			Reptile \cite{nichol2018first} & Computational Complexity & Standard stochastic gradient descent   & \CheckmarkBold & \XSolidBrush \\
			FOMAML \cite{nichol2018first} & Simplify secondary gradient updates & Using the gradient calculated from the previous task  & \CheckmarkBold & \XSolidBrush \\
			Meta-SGD \cite{li2017meta} & Increasing the Volume of the Model &  Increase the learning rate vector parameter & \XSolidBrush & \CheckmarkBold \\
			TAML \cite{jamal2019task} & Task unbiased estimation & Introducing Entropy and Inequality Metrics & \XSolidBrush & \CheckmarkBold \\
			iMAML \cite{rajeswaran2019meta} & Gradient disappearance & Propose new loss functions and optimization methods& \XSolidBrush & \CheckmarkBold \\
			iTMAML \cite{rajasegaran2020itaml}& Automatic task identification & Data is in continuous state & \XSolidBrush & \CheckmarkBold \\
			\bottomrule
		\end{tabular}
	}
\end{table*}

Learning optimizers are another important direction for learning model parameters. LSTM as the base optimizer \cite{wichrowska2017learned,chandriah2021rnn}, which accepts the difference at time $t$ and the hidden state of the meta-network at time $t-1$. The output of the original network is an updating of the model's weight and bias. In 2016, Xu et al. \cite{andrychowicz2016learning} proposed the BPTT to supervise LSTM training. It is notable that this is performed in the context of supervised learning. What update should be required to the optimization if it is in the setting of unsupervised and active learning? Inspired by this, there has been a long period of work focusing on reinforcement learning \cite{houthooft2018evolved}, Bayesian inference \cite{zintgraf2019varibad} and evolutionary algorithms \cite{piergiovanni2020evolving} in an attempt to automatically find optimization strategies through heuristic algorithms.

Finally, traditional Neural Architecture Search (NAS) also incorporated the idea of meta-learning and adapted it accordingly under FSL. To our best knowledge, the shared \cite{pham2018efficient} and randomly selected supernet weights \cite{liu2018darts,bender2018understanding} were early solutions for FSL. Recently, a large volume of work \cite{bender2018understanding,yu2019evaluating,dong2020bench} has shown that performance differences still exist between one-shot NAS and traditional NAS. The one-shot NAS uses weight-sharing networks to train the supernetwork only once and then perform a single round of inference to get an accurate prediction, greatly reducing the amount of computation required for the experiment. Subsequently, Zhao et al. \cite{zhao2021few} proposed the few-shot NAS based on one-shot NAS. The core idea is to divide the supernets into multiple sub-supernets to search different regions of the search space. With a slight increase in the number of supernets, the accuracy of few-shot NAS is greatly improved. MetaNAS \cite{elsken2020meta} is the first method that completely integrates meta-learning and traditional NAS. MetaNAS be capable of better initialization parameters with the help of meta-learning ideas. It completely replaces the weighted summation in the DARTS algorithm to reduce different operations, and the experimental results also show that it is more adaptable to more downstream learning tasks.

\subsection{Learning Metric Algorithm }
\begin{figure}[h]
	\centering
	\includegraphics[width=\linewidth]{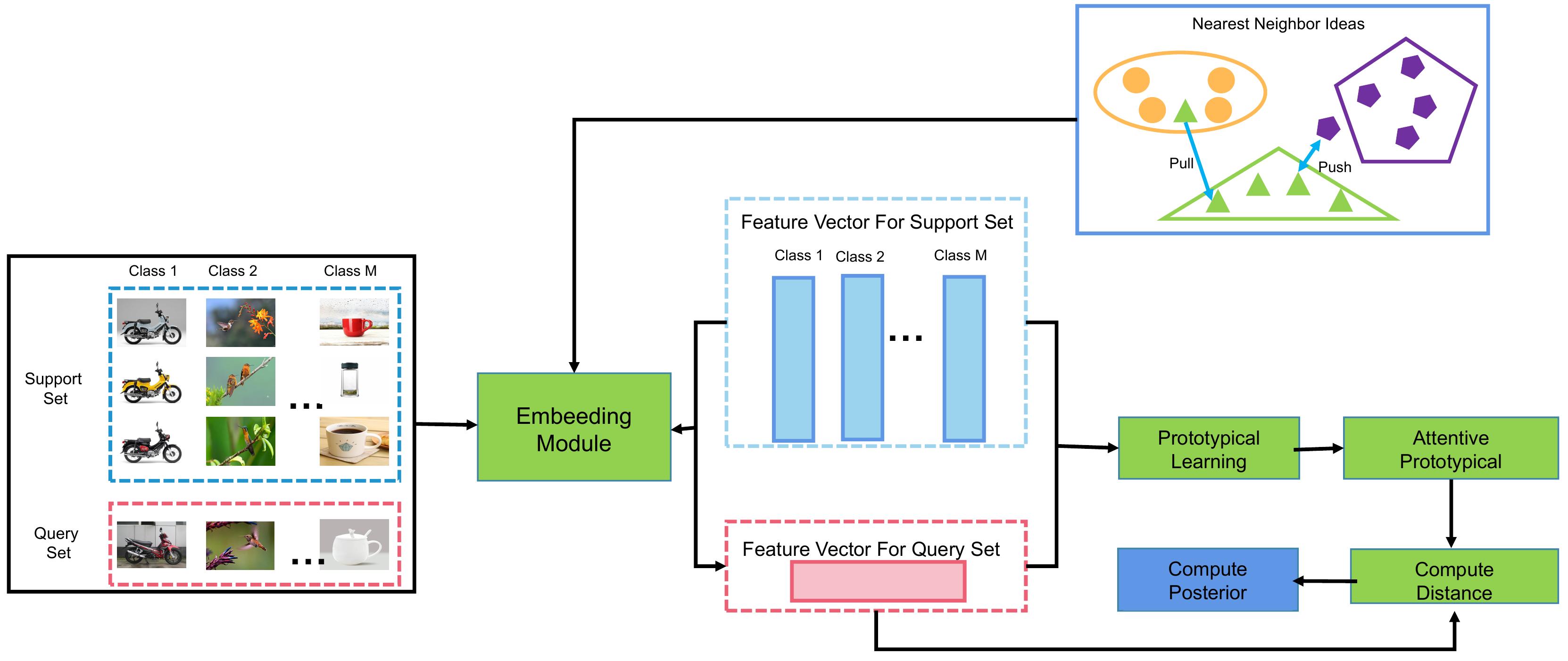}
	\caption{The framework \cite{wu2020attentive} performs end-to-end learning of the embedding model and prototype learning jointly, and the learned embedding features are used to compute the distance between the query image and the prototype, pushing the distance between different classes farther and bringing the distance between the same classes closer. }
	\label{newmetric}
\end{figure}

Metric learning \cite{musgrave2020metric} is different from classical meta-learning, metric learning no longer divides the model into training and testing stages. In many previous papers \cite{lu2020learning,wang2020generalizing,shu2018small,bendre2020learning}, metric learning is always introduced separately. In our context, metric learning will be explained under the frame of meta-learning. Fig. \ref{newmetric} illustrates one of the most representative learning methods, which is based on a prototype network that has been improved to obtain substantial improvements on a benchmark dataset for classification tasks. 

The siamese neural network \cite{koch2015siamese} is a relatively early model in the metric learning. It can be simply regarded as a binary classification problem. The input to the model composes of a set of positive or negative sample pairs, and the model needs to evaluate the similarity of the images during inference stage. Triple loss \cite{hoffer2015deep} is another way to deal with more than pairs input in FSL metric learning. 
Contrary to the Siamese neural network, triple loss requires positive samples, negative samples, and anchor samples to be available at the same time. If training samples are easily distinguished from each other, this would not be beneficial for the model to better learn discriminative features. The hard sample selection technique \cite{chen2017beyond} incorporates the absolute distance between positive sample pairs in addition to considering the relative distance between positive and negative samples. In addition to this, Li et al. \cite{li2020revisiting} revisited the classical triplet network and extended it to a K-tuple network for FSL.

Compared to the Siamese neural network, the prototype network \cite{snell2017prototypical} realizes the true meaning of classification. The most significant difference is that the model allows for more data as input. By feature averaging it is feasible to find the most representative sample as a prototype. However, simple feature averaging is easily disturbed by noise. On this basis, many works \cite{yang2006distance,schultz2004learning,chen2009similarity,weinberger2006distance,wu2020attentive} have explored how to make the distance between prototypes larger and larger. One of the most representative works is the proposal of positive and negative margins \cite{liu2020negative}, which further reduce the over-fitting and enhance the generalization based on maximizing the discriminative ability of the model.

Matching Networks \cite{vinyals2016matching} is a more general network framework that maps few-shot datasets and unlabeled data to vectors in the embedding space. The matching network combines the best features of parametric and nonparametric models of the nearest neighbour algorithm to model the sample distance distribution by learning the embedding representation. Experiments have proved \cite{rodriguez2020embedding} that embedding propagation produces a smoother embedding manifold. How to learn high-quality embedding representations in a limited time is substantial for improving the model's accuracy. GVSE \cite{huang2019few} fuses visual embeddings, semantic embeddings, and gating metrics automatically balances the relative importance of each metric dimension by the model. Subsequently, Arvind Srinivasan et al. \cite{srinivasan2020optimization} proposed a new architecture to improve Inception-Net, U-Net, Attention U-Net, and Squeeze-Net, which takes the time to generate embedding quality as a cost. The processing based on the embedding representation plays a vital role in FSL.

The relational network \cite{sung2018learning} differs from the three models mentioned above in that its similarity is calculated by using a neural network. In contrast to the Siamese neural networks and prototype networks, relational networks can be seen as providing a learnable nonlinear classifier for determining relationships. The classifier can be a feature extractor of a pre-trained neural network \cite{li2020revisiting} or a multiple embedded module \cite{li2020bsnet}. The most significant contribution of the relational network is that it breaks away from a single linear metric function and explores the use of an alternative model to generate similarity. Table. \ref{tab:learn1} categorizes each of the representative metric learning algorithms, comparing their innovations on the original approach.

\begin{table*}
	\centering
	\caption{A Summary of Metric Learning by base approach.
}
	\label{tab:learn1}
	\resizebox{\linewidth}{!}{
		\begin{tabular}{llll}
			\toprule
			Model &  Key idea & Metric function & Improvement \\
			\midrule
			\textbf{Siamese Neural Network \cite{koch2015siamese}} & \textbf{A pair of inputs} & \textbf{Cosine} & \textbf{Original}\\
			Triple loss \cite{hoffer2015deep} & Triple input & Cosine &\begin{tabular}[c]{@{}l@{}} Maximize intra-class distance and \\minimize inter-class distance\end{tabular}  \\
			E-nagivate sample \cite{chen2017beyond} & Difficult sample training & Cosine & Compare more samples at once\\
			K-tuple network \cite{li2020revisiting}& K inputs & Cosine & Compare more samples at once  \\
			\midrule
				\textbf{Prototype Network \cite{snell2017prototypical}} &  \textbf{Prototype representation}& \textbf{European}& \textbf{Original} \\
			Negative Margin Matters \cite{liu2020negative} & Negative margin loss& Cosine &Balancing discriminative and migratory \\
			Attentive Prototype \cite{wu2020attentive}  & \begin{tabular}[c]{@{}l@{}}Consider spatial association \\ between features\end{tabular} & European & Weighted summation to obtain the prototype \\
			SEN \cite{nguyen2020sen} & Feature normalization & European & \begin{tabular}[c]{@{}l@{}}The modal length of the constraint feature\\ approximates the modal length of the prototype \end{tabular}\\
			Prototype Rectification \cite{liu2020prototype}  & \begin{tabular}[c]{@{}l@{}}Modifying prototypes with \\ query sets\end{tabular}  & Cosine & Consistent distribution of query sets and support sets\\
			\midrule
				\textbf{Matching Network \cite{vinyals2016matching}} & \begin{tabular}[c]{@{}l@{}}\textbf{Attention mechanism to} \\ \textbf{access the memory matrix} \end{tabular} & \textbf{Cosine} &\textbf{Original} \\
			GVSE \cite{huang2019few}& \begin{tabular}[c]{@{}l@{}}The relative importance of the \\ automatic balancing model for \\each metric\end{tabular} & Cosine & \begin{tabular}[c]{@{}l@{}}Fuses visual embeddings, semantic embeddings, \\and gating metrics\end{tabular}\\
			Optimization of image embeddings \cite{srinivasan2020optimization}&Monitoring embedded quality & Manhattan & Improved quality of embedding\\
			\midrule
				\textbf{Relational Network \cite{sung2018learning}} & \textbf{Using models to compute similarity} & \textbf{Model} & \textbf{Original}  \\
			Revisiting metric learning \cite{li2020revisiting} & Using a simple but powerful baseline & non-linear distance & Proposing a deep K-tuplet network \\
			BSNet \cite{li2020bsnet}& Simultaneously by two similarity measures & Euclidean and cosine & \begin{tabular}[c]{@{}l@{}}Learning feature maps based on the similarity \\of two different features\end{tabular}\\
			\bottomrule
		\end{tabular}
	}
\end{table*}

\subsection{Learning To Transmit Information } 
\begin{figure}[h]
	\centering
	\includegraphics[width=\linewidth]{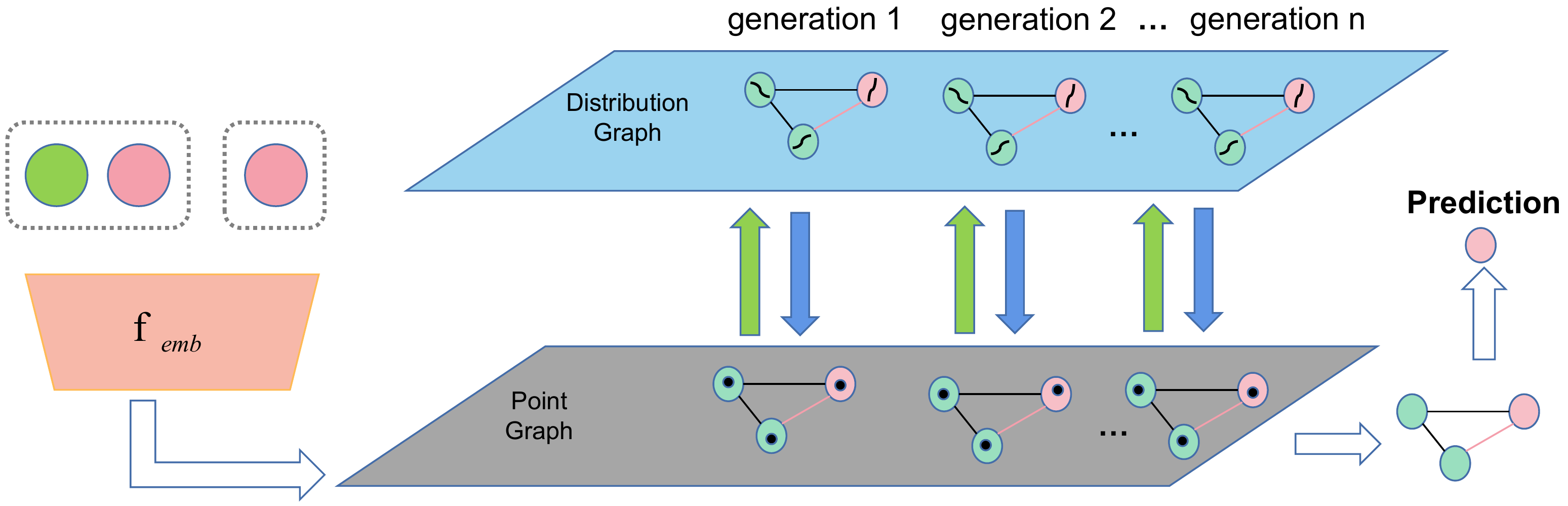}
	\caption{DPGN \cite{yang2020dpgn} is concerned with the relationship between samples in GNN, in addition to the relationship between sample distributions. Where Point Graph is used to describe the samples and Distribution Graph is used to describe the distribution. The two GNNs fuse the instance-level and distribution-level relationships by passing information. }
	\label{graph}
\end{figure}

It is proved that graph neural networks (GNNs) \cite{wu2020comprehensive} have performed well on relational-based tasks in recent years \cite{garcia2017few}. Researchers have found that its classes-based transfer of information can work well to help FSL learn to identify new class, while avoiding these classes being dominated by proprietary features. Primarily, early graph neural networks simulate the propagation of weights between different nodes by creating full connections between support and query sets. The nodes can be represented by either a one-hot encoding or an embedding vector, and the connections between nodes can be passed through edges. Given the complexity of graph neural network algorithms, most graph neural networks currently have a shallow number of layers. In order to better accommodate FSL, graph neural networks have been uniquely designed with nodes and edges in recent developments. Fig. \ref{graph} shows a recent representative algorithm for FSL of graph neural networks from the perspective of exploring small sample distributions.

The EGNN uses vertex sets, edge sets, and task sets to encode the labels of nodes. When updates occur between nodes, both the similarity and the difference are considered, which greatly improves the generalization performance of graph neural networks to FSL. Meta-GCN \cite{bose2019meta} further incorporates the idea of meta-learning, which enables the updating of weights of graphs under FSL to also be optimized according to the gradient descent steps, the whole process requires very few gradient steps and can receive new data quickly. Subsequently, several models based on improvements in the graph structure itself have emerged. The prototype network was improved by GFL \cite{yao2020graph} network that focuses on learning small samples of data with a graph structure. DPGN uses a dual graph neural network that describes samples while modelling their distribution. Furthermore, GERN \cite{liu2020graph} uses embedding of graph neural network connections to achieve more robust intra-class weight transfer. Nevertheless, none of these approaches addressed the problem of shallow layers of graph neural networks until 2021, HGNN \cite{chen2021hierarchical} designed three sections of bottom-to-top, and skip connections to remove the pitfall of ordinary GNNs losing the hierarchical association between nodes. Based on this, Frog-GNN \cite{xu2021frog} uses multidimensional information to synthesize information about the adjacency between nodes to form pairwise relational features of intra-class similarity and inter-class dissimilarity. At present, graph neural networks are widely used for tasks such as few-shot image classification \cite{gidaris2019generating, xiong2021multi}, semantic segmentation \cite{xie2021scale} and instance segmentation tasks.
\subsection{Discussion and Summary}
In FSL, meta-learning mainly explore the mapping from the task to the target model. It trains a super-tuning device that gives a good set of hyperparameters as it converges according to the different tasks. In contrast to multi-task learning, which learns only focus on single task. However, meta-learning is not universal for all conditions. The current idea of meta-learning is to have enough historical tasks. If there are not enough tasks on certain problems, then meta-learning may not be able to solve those problems. Similarly, if the domain gap between source and target is too large, the results  will also become terrible.

\section{Multimodal Complementary Learning Of Small Samples With Limited Information }
\label{secmul}
\begin{figure}[h]
	\centering
	\includegraphics[width=\linewidth]{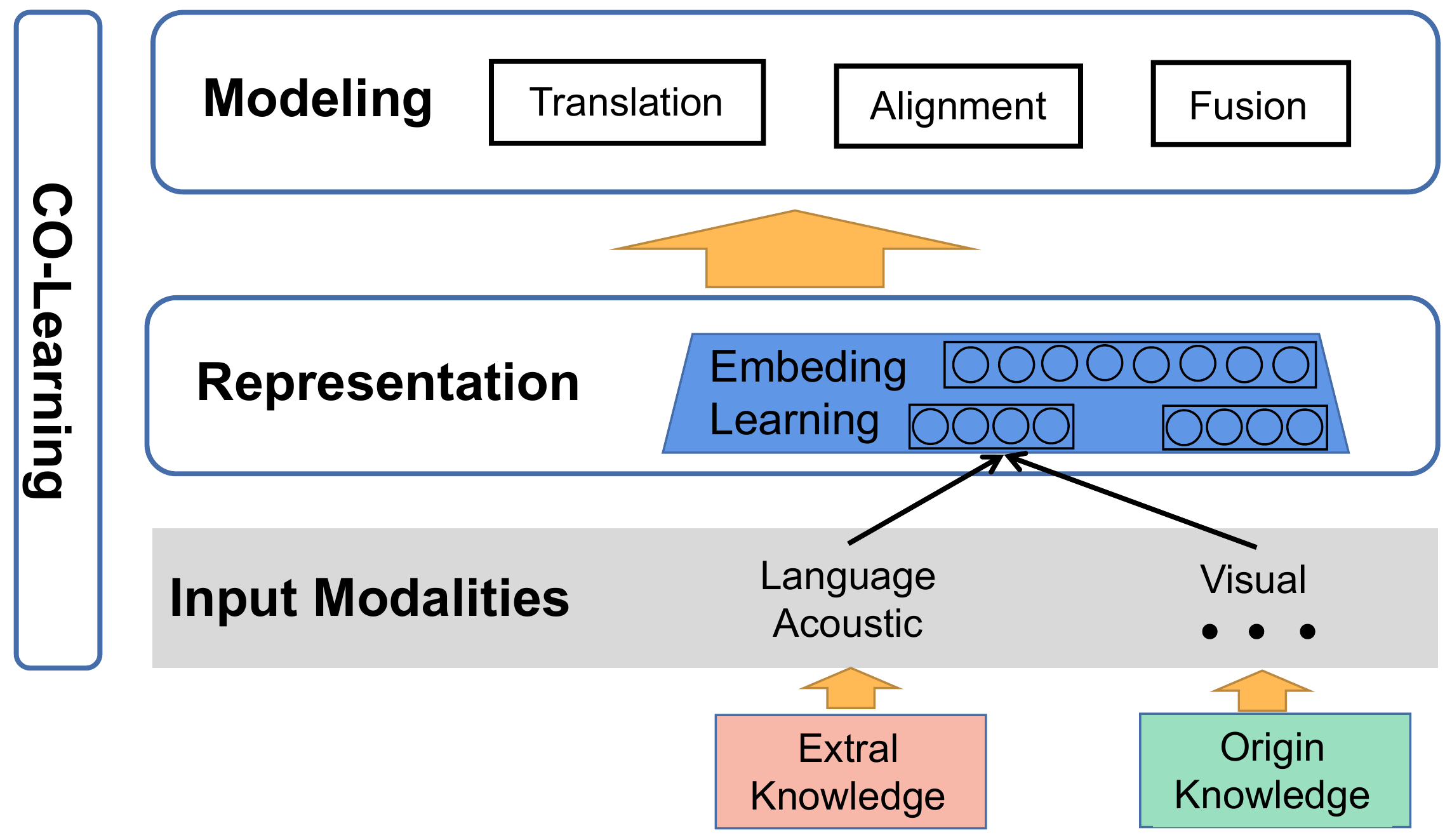}
	\caption{Multimodal FSL scenarios, how to effectively model other modal information under the condition of feature representation by fusion, alignment, and assistance to compensate for the lack of valid information in itself.}
	\label{colearning}
\end{figure}

Until now, FSL has made significant progress in the unimodal domain. Within unimodal learning, models are primarily responsible for representing information as feature vectors that can be processed by a computer or further abstracted into higher-level semantic vectors. Particularly, multimodal learning in FSL refers to learning better feature representations by exploiting complementarities between multiple modalities and removing redundancies between modalities. In real life, when parents teach their babies about things, they always include general information along with semantic descriptions. This is crucial for FSL, which inherently comes with little valid information to make a good evaluation of the data or feature distribution. Inspired by this, many research works \cite{wang2020large,li2019large,schonfeld2019generalized} consider the introduction of other modal information when solving FSL. By fusing multimodal information, the ability of the model to perceive small sample data can be improved. Fig. \ref{colearning} shows the main paths of FSL under multimodality, 

\begin{table}
	\centering
	\caption{The challenge of learning from small samples in multimodality.}
	\label{tab:learn3}
	\resizebox{\linewidth}{!}{
    \small
		\begin{tabular}{lccccc}
			\toprule
			Approach &  Representation & Alignment  & Fusion &  Co-Learning & Translation  \\
			\midrule
			 Wang et al. \cite{wang2020large} & \CheckmarkBold & & & &\\
			Li et al. \cite{li2019large} &  & & &\CheckmarkBold &\\
			Eli et al. \cite{schwartz2019baby} &  & & \CheckmarkBold & &\\
			Peng et al. \cite{peng2019few} & \CheckmarkBold & & & \\
			Schonfeld et al. \cite{schonfeld2019generalized} & &\CheckmarkBold  & & \\
			Wang et al. \cite{wang2019tafe} & & \CheckmarkBold & &\\
			Pade et al. \cite{pahde2021multimodal} & \CheckmarkBold & & &\\
			Fortin et al. \cite{fortin2019multimodal} & & & &\CheckmarkBold & \\
			Zhang et al. \cite{zhang2017stackgan} & \CheckmarkBold & & & &\\
			Sharma et al. \cite{sharma2018chatpainter} & & & & & \CheckmarkBold\\
			Akata et al. \cite{reed2016learning} &\CheckmarkBold & & & &\\
			Elhoseiny et al. \cite{elhoseiny2017link} & & & & &\CheckmarkBold\\
			Zhu et al. \cite{zhu2018generative}& & & & \CheckmarkBold&\\
			Xian et al. \cite{xian2018feature}& & &\CheckmarkBold & &\\
			Pahde et al. \cite{pahde2021multimodal}& & & & &\CheckmarkBold\\
			\bottomrule
		\end{tabular}
	}
\end{table}
\subsection{Multimodal embedding}
Recent works \cite{xing2019adaptive,li2019large,schonfeld2019generalized,schwartz2019baby,karpathy2015deep} proved the limitations of visual features for FSL of certain tasks. Semantic space as auxiliary information can provide effective context for visual features and help FSL. Experiments have shown \cite{wang2020large,li2019large,schonfeld2019generalized} that adaptive combinations of two or more modalities are much better than unimodal FSL. Wang et al. \cite{wang2020large} constructed weak semantic supervision for each category by integrating multiple visual features. Schonfeld et al. \cite{schonfeld2019generalized} instead used variational autoencoders (VAEs) to model semantic features based on latent visual features. Subsequently, Schwartz et al. \cite{schwartz2019baby} and Peng et al. \cite{peng2019few} further extended the semantic information by adding classes labels, attribute and natural language descriptions, and knowledge inference. The additional semantic information is aligned with visual features by embedding loss functions \cite{wang2019tafe} to largely reduce the cost of knowledge transfer.
Based on this, Karpathy et al. \cite{karpathy2015deep} used multimodal alignment to find potential correspondences that exist between image patches in the training set images and their descriptive utterances. Aoxue et al. \cite{li2019large} went further by using semantic information to model classes as hierarchy.

\subsection{Generate semantic information from images}
In addition to this, another related area using multimodal FSL is text-to-image generation. In few-shot visual classification tasks, the visual and semantic-based approaches \cite{pahde2021multimodal}, which try to use textual descriptions to generate additional training images, have a considerable advantage. Pade et al. \cite{pahde2021multimodal} used generative adversarial networks as data generators to train the model, which can purposefully generate corresponding visual features based on semantic information, and enhanced visual features can be obtained by combining the original visual features. Zhu et al. \cite{zhu2018generative} and Xian et al. \cite{xian2018feature} explored generative images and feature vectors, respectively, making promising progress in the field of ZSL.

%to assist zero-shot learning?

Similarly, Fortin et al. \cite{fortin2019multimodal} migrated text-to-image generation to the target detection task, which can be integrated with current FSL to implement a more general module in the contextual joint learning phase. Zhang et al. \cite{ zhang2017stackgan} improved the resolution of the generated images by concatenating two CGANs based on ordinary generative networks. The first subtask generates a relatively blurred image from the text, and the second subtask generates a high resolution image from the blurred image. Eventually, the model will use more details to generate images. Nevertheless, sometimes text descriptions contain multiple targets and a single text description does not capture all the details in an image. Sharma et al. \cite{sharma2018chatpainter} provided a dialogue interface that uses textual information from the dialogue to obtain more detailed information about the image.
 
Another set of text-to-image algorithms is based on the variant auto-encoder with embedding. Unlike the generative approach, the input to the encoder is a vector of attributes. Akata et al. \cite{reed2016learning} explored semantic features from different sources, such as WordNet and word embeddings. However, these methods were unable to recognize parts of an image without part-term annotation. Elhoseiny et al. \cite{elhoseiny2017link} used a visual classifier to detect patches from bird image dataset by using only text terms and tests without partial annotation. The results show that visual text information and bird parts can be linked with zero samples.

\subsection{Discussion and Summary}
Multi-modal FSL is still in the developing stage and there are currently several challenges until now: how to combine data from heterogeneous domains, how to deal with the different levels of noise that occur during the combination of different modalities, and how to learn together. Table. \ref{tab:learn3} classifies FSL tasks in multimodality as representation, alignment, fusion, co-learning, and translation. In a multimodal FSL, a good feature representation should be able to fill in the missing modalities based on the observed modal information. More approaches will emerge in the future, going well beyond modal embedding and generating semantic information from images.

\section{FSL Applications In Computer Vision}
\label{seccv}
In the past five years, we systematically combed and summarized FSL in the field of computer vision \cite{bateni2020improved}  and divided tasks into image classification, object detection, semantic segmentation, and instance segmentation. Following is a detailed summary in the form of graphs and tables based on the time dimension. By reading this section, the reader will be able to gain a comprehensive grasp of FSL in the field of computer vision.

\subsection{Few-shot Image Classification}

\begin{figure*}[h]
	\centering
	\begin{minipage}[t]{0.45\linewidth}
	\centering
	\includegraphics[width=\linewidth]{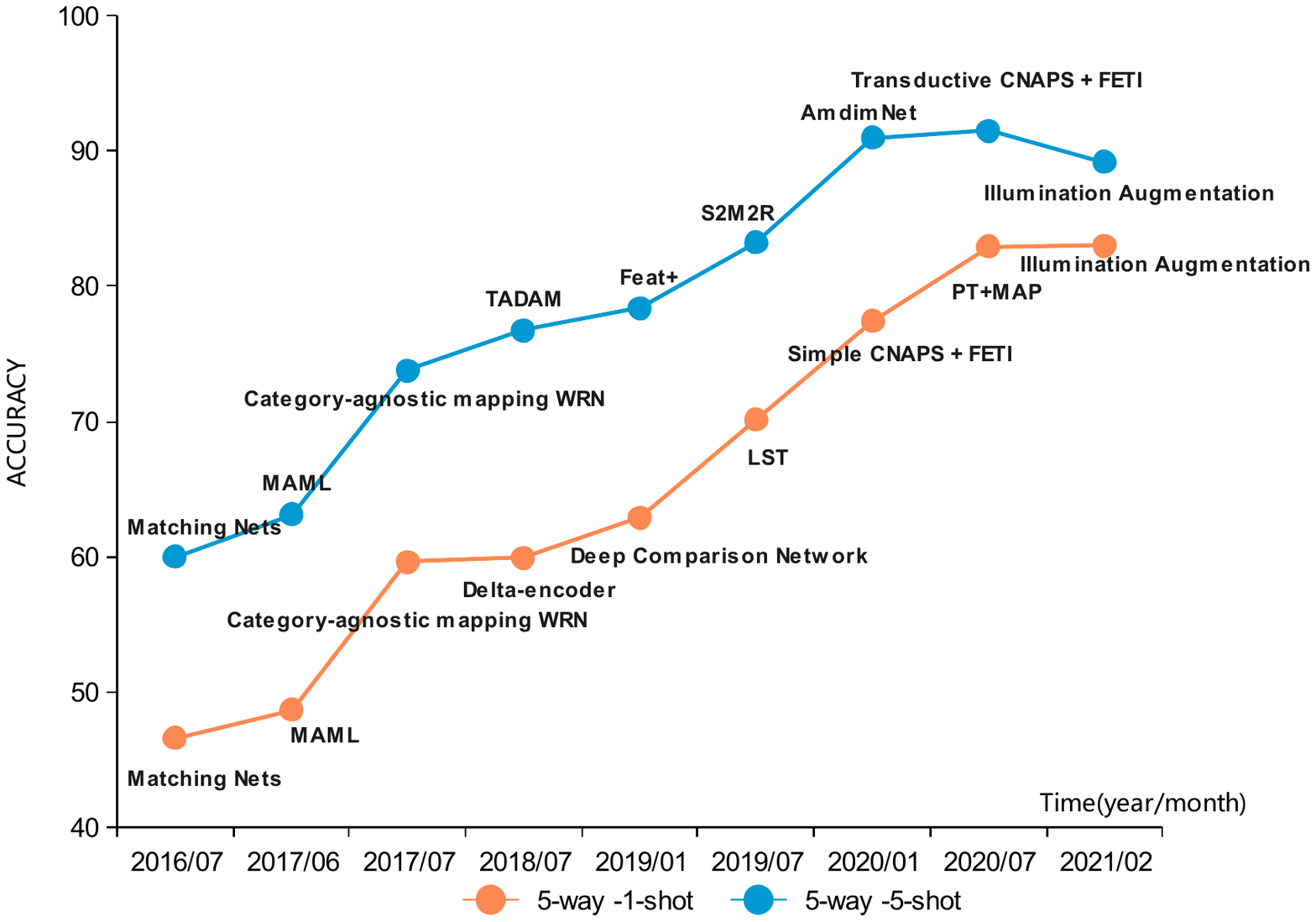}
	\caption{Best performance of metric learning in image classification tasks during 2017-2021}
	\label{image}
	\end{minipage}
	\hspace{0.35cm}
	\begin{minipage}[t]{0.45\linewidth}
		\centering
		\includegraphics[width=0.7\linewidth]{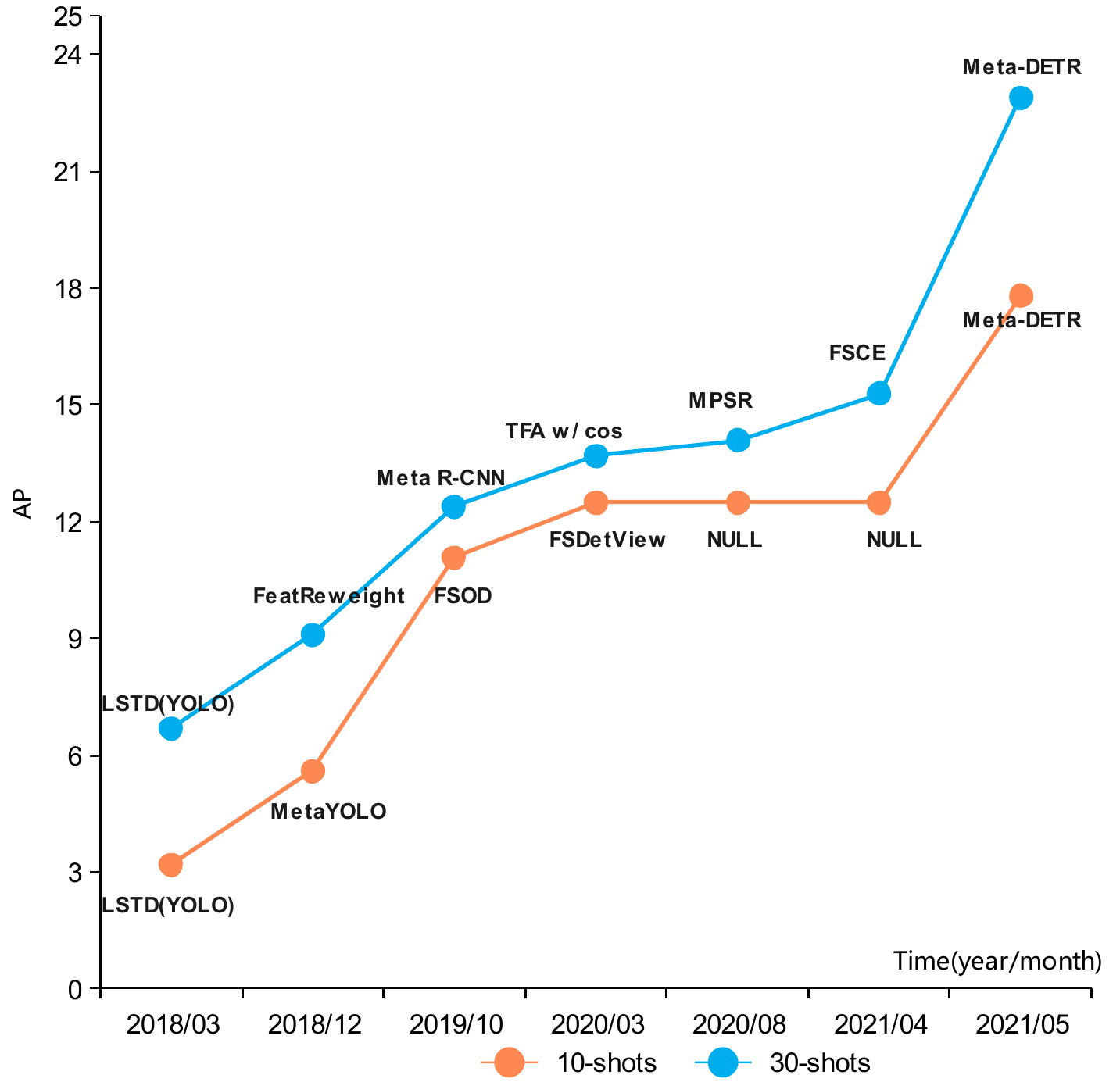}
		\caption{Best performance of metric learning in object detection tasks during 2017-2021}
	\end{minipage}
\end{figure*}

Except like Google and Facebook, most researchers in real life do not have access to a large dataset of good quality. In FSL computer vision classification tasks, each task may contain only one or a few samples. Solving few-shot image classification tasks is mainly addressed by data augmentation, transfer learning, meta-learning, and multimodal fusion learning. At present, the top three methods in terms of accuracy are all based on feature augmentation and feature transformation of the backbone model. In this section, we investigates all few-shot image classification models from 2016 to the present and counts the best performance of all models on the mini-ImageNet benchmark dataset. Here we use 5-way-1-shot and 5-way-5-shot as baseline tasks. Table. \ref{tabImage} and Fig. \ref{image} illustrate our investigation results. 

\begin{table*}
	\caption{The latest performance of FSL in the image classification tasks of computer vision}
	\label{tabImage}
	\resizebox{\linewidth}{!}{
		\begin{tabular}{llccclcc}
			\toprule
			Ref & Model & 5-way-1-shot accuary & 5-way-5-shot accuary & Extra Training Data & Approach Features & Pubilsed Date & Available code\\
			\midrule
			 Lee et al. \cite{lee2021unsupervised} & 	
			 ESFR & 76.84\% & 84.36\% &  \CheckmarkBold &\begin{tabular}[c]{@{}l@{}}Early-Stage Feature Reconst\\-ruction  \end{tabular}  & Jun 2021 & \CheckmarkBold \\
			Wang et al. \cite{wang2021bridging} & 	
			Multi-Task Learning & 59.84\% & 77.72\% & \XSolidBrush & \begin{tabular}[c]{@{}l@{}}Multi-Task Learning and Meta\\-Learning \end{tabular}  & Jun 2021 & \CheckmarkBold\\
			Afham et al. \cite{afham2021rich} & 	
			RS-FSL & 65.33\% & - & \CheckmarkBold & Semantic assistance & Apr 2021 & \CheckmarkBold\\
			 Rizve et al. \cite{rizve2021exploring} & 	
			Invariance-Equivariance & 67.28\% & 84.78\% & \XSolidBrush &\begin{tabular}[c]{@{}l@{}}Invariant and Equivariant Repres\\-entations\end{tabular}& Apr 2021 & \CheckmarkBold\\
			Esfandiarpoor et al. \cite{esfandiarpoor2020extended}& 	
			pseudo-shots & 73.35\% & 82.51\% & \XSolidBrush & Exploiting Existing Resources & Dec 2020 & \CheckmarkBold\\
			Chen et al. \cite{chen2020multi}& 	
			MATANet & 53.63\% & 72.67\% & \XSolidBrush & Multi-scale Adaptive + Attention & Nov 2020\\
			Wang et al. \cite{wang2020match} & 	
			MTUNet & 55.03\% & 56.12\% & \XSolidBrush &\begin{tabular}[c]{@{}l@{}}using backbone model and weight\\ generated\end{tabular}   & Nov 2020  & \CheckmarkBold \\
			Khacef et al. \cite{khacef2020gpu} & 	
			WRN + Self-Organizing Map & 71.5\% & 82.2\% & \XSolidBrush & Self-Organizing Maps & Sep 2020 &\XSolidBrush  \\
			Xue et al. \cite{xue2020region} & 	
			RCN - ResNet12 & 57.40\% & 75.19\% & \XSolidBrush & Transfer learning & Sep 2020 & \CheckmarkBold  \\
			Zhong et al. \cite{zhong2021complementing}& 	
			MCRNET & 62.53\% & 	80.34\% & \XSolidBrush & Meta-learning & Jul 2020 & \CheckmarkBold  \\
			Ziko et al. \cite{ziko2020laplacian} & 		
			LaplacianShot & 75.57\% & 84.72\% & \XSolidBrush & Transductive Laplacian-regularized & Jul 2020 & \CheckmarkBold  \\
			 Bateni et al. \cite{bateni2020enhancing} & 	
			Transductive CNAPS + FETI & 79.9\% & 91.5\% & \CheckmarkBold & Data Augmentation & Jun 2020  & \XSolidBrush  \\
			Rajasegaran et al. \cite{rajasegaran2020self} & 	
			SKD & 67.04\% & 83.54\% & \XSolidBrush & Knowledge Distillation & Jun 2020 & \CheckmarkBold  \\
			Hu et al. \cite{hu2020leveraging} & 	
			PT+MAP & 82.92\% & 88.82\% & \XSolidBrush & 	
			Feature Distribution & Jun 2020  & \CheckmarkBold  \\
			Simon et al. \cite{simon2020adaptive} & 	
			Adaptive Subspace Network & 67.09\% & 81.65\% & \XSolidBrush & central block of a dynamic classifier & Jun 2020 & \CheckmarkBold  \\
			Bateni et al. \cite{bateni2020enhancing} & 	
			Transductive CNAPS & 55.6\% & 73.1\% & \XSolidBrush & Self-Organizing Maps & Jun 2020 & \CheckmarkBold   \\
			Li et al. \cite{li2020boosting} & 	
			TRAML & 67.10\% & 79.54\% & \XSolidBrush & Adaptive Margin Loss & May 2020 & \CheckmarkBold   \\
			Hu et al. \cite{hu2020empirical} & 	
			SIB & 70.0\% & 79.2\% & \XSolidBrush & Empirical Bayes Transductive  & May 2020 & \CheckmarkBold   \\
			Wang et al. \cite{wang2020instance}& 	
			ICI & 69.66\% & 80.11\% & \XSolidBrush & Instance Credibility Inference & Apr 2020 &\CheckmarkBold  \\
			Rodríguez et al. \cite{rodriguez2020embedding}& 	
			EPNet + SSL & - & 88.05\% & \CheckmarkBold & Embedding Propagation & Apr 2020 & \CheckmarkBold  \\
			Nguyen et al. \cite{nguyen2020pac} & 	
			SImPa & 52.11\% & 63.87\% & \XSolidBrush & PAC-Bayes framework  & Mar 2020 & \XSolidBrush  \\
			Guan et al. \cite{guan2020few}& 	
			DAPNA & 71.88\% & 84.07\% & \CheckmarkBold & Domain Adaptation & Feb 2020 & \XSolidBrush  \\
			Chen et al. \cite{chen2021self} & 	
			AmdimNet & 	76.82\% & 90.98\% & \CheckmarkBold & Embedding network  & Nov 2019 & \CheckmarkBold  \\
			Xu et al. \cite{xu2020metafun} & 	
			MetaFun-Attention & 64.13\% & 80.82\% & \XSolidBrush & \begin{tabular}[c]{@{}l@{}}Use similar gradient descent \\to encode labeled data \\to predict unlabeled data.  \end{tabular}& Jau 2020 & \CheckmarkBold  \\
			Liu et al. \cite{liu2020task} & 	
			MetaOptNet-SVM+Task Aug & 	65.38\% & 82.13 \% & \CheckmarkBold & Embedding Propagation & Nov 2019 & \CheckmarkBold  \\
			Song et al. \cite{song2019generalized} & 	
			ACC + Amphibian & 62.21\% & 80.75\% & \CheckmarkBold & \begin{tabular}[c]{@{}l@{}}Uing pre-trained base model \\to generalize novel model \end{tabular}& Nov 2019 & \CheckmarkBold  \\
			Rodríguez et al. \cite{patacchiola2020bayesian}& 	
			DKT + BNCosSim & 62.96\% & 64\% & \XSolidBrush & \begin{tabular}[c]{@{}l@{}}Learn a kernel that\\ transfers to new tasks \end{tabular} & Dec 2019 & \CheckmarkBold  \\
			Mangla et al. \cite{mangla2020charting}& 	
			S2M2R & 64.93\% & 83.18\% & \CheckmarkBold & Embedding Propagation & Apr 2020 & \CheckmarkBold  \\
			Li et al. \cite{li2019learning} & 	
			LST & 70.1\% & 78.7\% & \CheckmarkBold & \begin{tabular}[c]{@{}l@{}}Train a small sample\\ model to predict fake \\signatures on unlabeled data  \end{tabular} & Sep 2019 & \CheckmarkBold  \\
			Yoon et al. \cite{yoon2019tapnet}& 	
			TapNet & 61.65\% & 76.36\% & \CheckmarkBold & \begin{tabular}[c]{@{}l@{}}Learn the reference vector \\of each class in \\different tasks  \end{tabular}  & May 2019 & \CheckmarkBold  \\
			Kim et al. \cite{kim2019edge} & 		
			EGNN + Transduction & - & 76.37\% & \XSolidBrush & \begin{tabular}[c]{@{}l@{}}Using graph neural network \\to model intra-class similarity  \end{tabular}    & May 2019 & \CheckmarkBold  \\
			Ye et al.\cite{ye2020few} & 	
			feat+ & 61.72\% & 78.38\% & \XSolidBrush & \begin{tabular}[c]{@{}l@{}}Set-to-set applied \\ to embedded functions  \end{tabular}& Jau 2019 & \CheckmarkBold  \\
			Li et al. \cite{li2019revisiting} & 	
			DN4 & 51.24\% & 71.02\% & \XSolidBrush & \begin{tabular}[c]{@{}l@{}}Use image local descriptors\\ for measurement  \end{tabular}   & Jun 2019 & \CheckmarkBold  \\
			Park et al. \cite{park2019meta}& 	
			MC2+ & 55.73\% & 70.33\% & \XSolidBrush & Factorization matrix   & Jau 2019 & \CheckmarkBold  \\
			\bottomrule

		\end{tabular}
	}
\end{table*}

\subsection{Few-shot Object Detection}

Few-Shot Object Detection (FSOD) is the task of detecting rare objects from several samples. There has been a lot of progress in FSL for image classification, but rarely for object detection. At present the evolution of few-shot object detection can be divided into three main camps: Data augmentation, transfer learning, and meta-learning. Out of them, Attention mechanisms plays a pivotal role in small sample target detection. Equally, the issue of slow inference for a few-shot object detections to meet real-time requirements remains serious. The Table. \ref{tab:object} and Fig. \ref{image} are used to show recent advances in object detection in FSL.

\begin{figure*}[h]
	\centering
	\begin{minipage}[t]{0.4\linewidth}
		\centering
		\includegraphics[width=0.75\linewidth]{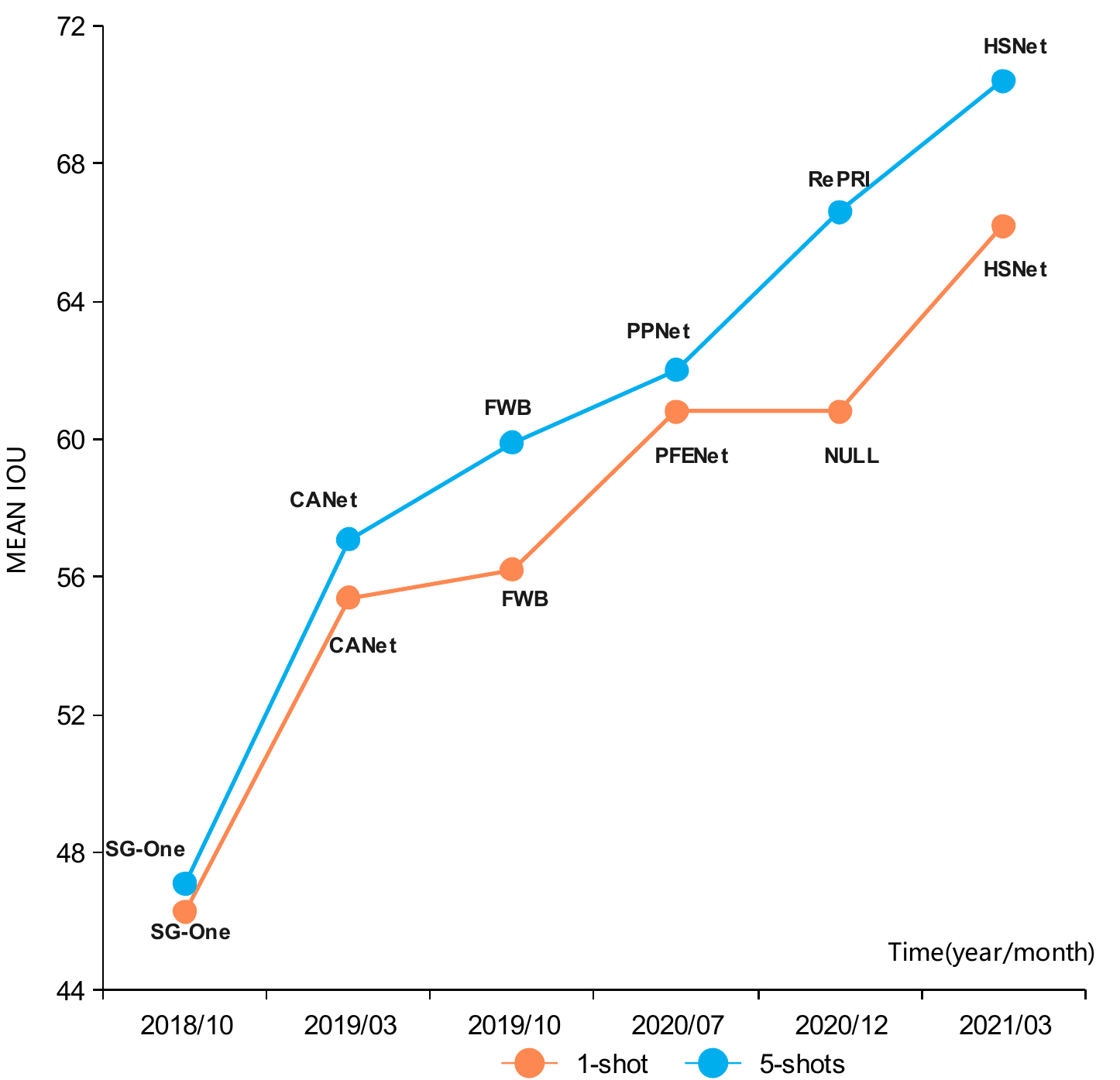}
		\caption{Best performance of metric learning in semantic segmentation tasks during 2018-2021}
		\label{semantic}
	\end{minipage}
	\hspace{2cm}
	\begin{minipage}[t]{0.4\linewidth}
		\centering
		\includegraphics[width=\linewidth]{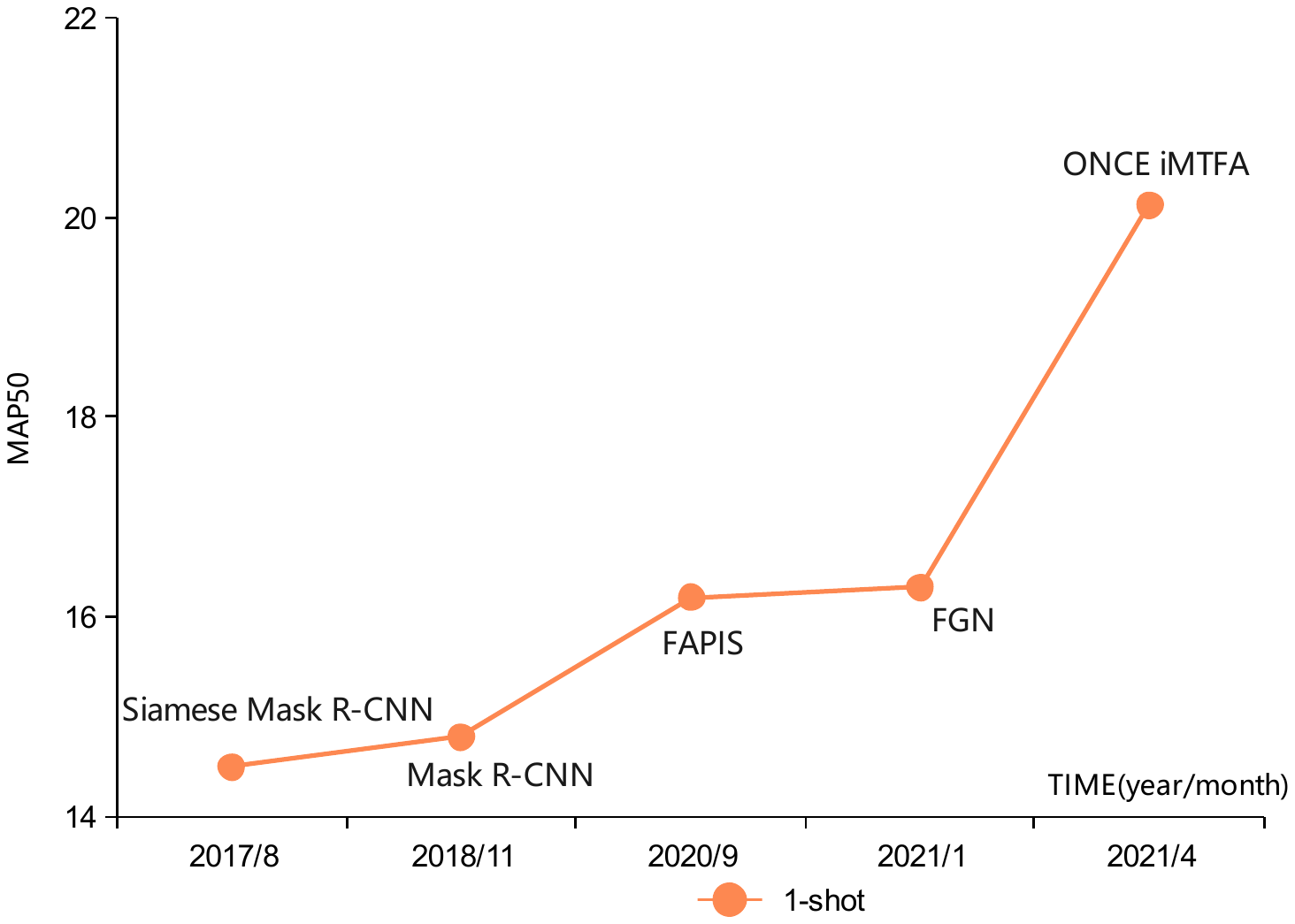}
		\caption{Best performance of metric learning in instance segmentation tasks during 2017-2021}
	\end{minipage}
\end{figure*} 

\begin{table*}
	\caption{The latest performance of FSL in the tasks of few-shot object detection during 2019-2021}
	\label{tab:object}
	\resizebox{\linewidth}{!}{
		\begin{tabular}{llccllcc}
			\toprule
			Ref & Model & \begin{tabular}[c]{@{}l@{}} 10-shot AP/\\30-shot AP \end{tabular} & Extra Training Data & Core idea & Approach Taxonomy & Pubilsed Date & Available code\\
			\midrule
			Zhang et al. \cite{zhang2021meta} & 	
			Meta-DETR & 17.8/22.9 & \XSolidBrush & \begin{tabular}[c]{@{}l@{}}Use semantic alignment to\\ perform specific encoding and\\ feature-independent decoding \\of images\end{tabular}  & Feature Reconstruction & Jun 2021 & \CheckmarkBold \\
			Sun et al. \cite{sun2021fsce} & 	
			FSCE & 11.1/15.3 & \XSolidBrush &\begin{tabular}[c]{@{}l@{}} Compare proposal coding loss \\to improve intra-class compactness\\ and inter-class variance \end{tabular} & Feature embedding & Jun 2021 & \CheckmarkBold \\
			Zhu et al. \cite{zhu2021semantic} & 	
			SSR-FSD & 11.3/14.7 & \CheckmarkBold & \begin{tabular}[c]{@{}l@{}} Learning Semantic Embeddings\\Using the Invariance of\\Semantic Relations \end{tabular} & Embedding learning & Mar 2021 & \XSolidBrush\\
			Xiao et al. \cite{xiao2020few} & 	
			FsDetView & 12.5/14.7 & \CheckmarkBold & \begin{tabular}[c]{@{}l@{}} Share the features of\\ the base class and \\the new class \end{tabular}  & Meta-learning & Jul 2020 & \CheckmarkBold\\
			Wu et al. \cite{wu2020multi}& 	
			MPSR & 9.8/14.1 & \XSolidBrush & \begin{tabular}[c]{@{}l@{}} Refinement of samples using \\multi-scale techniques \end{tabular} & Data augmentation & Jul 2020 & \CheckmarkBold\\
			Wang et al. \cite{wang2020frustratingly} & 	
			TFA w/ cos & 10/13.7 & \XSolidBrush & \begin{tabular}[c]{@{}l@{}} Fine-tuning the final\\ layer of the detector \end{tabular} & Fine-tuning & Mar 2020 & \CheckmarkBold\\
			Yan et al. \cite{yan2019meta} & 	
			Meta R-CNN & -/12.4 & \XSolidBrush &\begin{tabular}[c]{@{}l@{}} Meta-learning using partial \\features \end{tabular}  & Meta-learning & Sep 2019  & \CheckmarkBold \\
			Wang et al. \cite{wang2019meta} & 	
			MetaDet & 7.1/11.3  & \CheckmarkBold & \begin{tabular}[c]{@{}l@{}} Prediction of component-specific \\parameters from several samples \end{tabular} & Meta-learning & Sep 2020 & \XSolidBrush  \\
			\bottomrule
		\end{tabular}
	}
	\vspace{-0.4cm}
\end{table*}

\subsection{Few-shot Semantic Segmentation }

Few-shot semantic segmentation was first proposed in \cite{shaban2017one} until 2017. And it has been widely used in scenes such as medical images and driverless cars. Unlike traditional semantic segmentation, few-shot semantic segmentation has less pixel annotation information in support data set. To our best knowledge, few-shot semantic segmentation can be broadly classified into supervised semantic segmentation, unsupervised semantic segmentation, and video semantic segmentation. In the machine learning stage, the more classical approach is to use probabilistic mappings as prior knowledge for derivation. In the deep learning phase, a large number of efficient algorithms for segmentation tools have emerged, but these models often require a large number of manual sample annotations. Recently, \cite{lu2021simpler} has made significant improvements to few-shot semantic segmentation by proposing a more concise paradigm where only the classifier is meta-learned and the feature encoding decoder remains trained using a conventional segmentation model. Providing the Table. \ref{semantic} and Fig. \ref{image} for showing few-shot semantic segmentation.

\begin{table*}
	\caption{The latest performance of FSL in the tasks of few-shot semantic segmentation during 2019-2021}
	\label{tab:semantic}
	\resizebox{\linewidth}{!}{
		\begin{tabular}{llccllcc}
			\toprule
			Ref & Model & \begin{tabular}[c]{@{}l@{}} 1-shot Mean IoU/\\5-shot Mean IoU \end{tabular}& Extra Training Data & Core idea & Approach Taxonomy & Pubilsed Date & Available code\\
			\midrule
			Zhang et al. \cite{zhang2021few} & 	
			CyCTR & 64.3/66.6 & \XSolidBrush & \begin{tabular}[c]{@{}l@{}} Aggregate supported and pixel \\features into query sets \end{tabular}  & Transfer learning & Jun 2021 & \CheckmarkBold \\
			Min et al.\cite{min2021hypercorrelation} & 	
			HSNet & 66.2/70.4 & \CheckmarkBold & \begin{tabular}[c]{@{}l@{}} Extracting different sets of \\feature composition from different\\ levels of intermediate convolutional \\layers \end{tabular}  & Feature Engineering & Apr 2021 & \CheckmarkBold \\
			Yang et al. \cite{yang2020prototype} & 	
			RPMM  & 56.3/- & \XSolidBrush & \begin{tabular}[c]{@{}l@{}} Different image regions are \\associated with multiple prototypes \\to obtain a semantic\\ representation \end{tabular} & Metric learning & Sep 2020 & \CheckmarkBold\\
		    Tian et al. \cite{tian2020prior}& 	
			PFENet & 60.8/- & \XSolidBrush & Feature enrichment + a priori mask  & Feature Engineering & Aug 2020 & \CheckmarkBold\\
			Liu et al. \cite{liu2020part}& 	
			PPNet & 51.5/62.0 & \CheckmarkBold & \begin{tabular}[c]{@{}l@{}} Refinement of samples using \\multi-scale techniques \end{tabular} & GNN & Sep 2020 & \CheckmarkBold\\
			Nguyen et al. \cite{nguyen2019feature}& 	
			FWB  & 56.2/59.9  & \XSolidBrush &\begin{tabular}[c]{@{}l@{}} Measuring the cosine similarity\\ of class feature vectors \\and query feature vectors \end{tabular}  & Metric learning & Sep 2019 & \CheckmarkBold\\
			Wang et al. \cite{wang2019panet}& 	
			PANet & 48.1/55.7 & \XSolidBrush & Each pixel is compared to the prototype & Metric learning & Feb 2020  & \CheckmarkBold \\
			Zhang et al. \cite{zhang2019canet} & 	
			CANet & 55.4/57.1  & \XSolidBrush & \begin{tabular}[c]{@{}l@{}} Support for performing multi\\-level feature comparisons between \\images and query images \end{tabular} & Metric learning  & Mar 2019 & \CheckmarkBold  \\
			\bottomrule
		\end{tabular}
	}
\end{table*}

\subsection{Few-Shot Instance Segmentation}
In contrast to semantic segmentation, instance segmentation involves identifying each pixel in an image and labelling it separately. Recently, few studies are dealing with the problem of segmenting few samples of instances. Current work still focuses on how to improve R-CNNs using some effective tools. The most recent work \cite{ganea2021incremental} proposes an incremental few-shot instance segmentation algorithm, which greatly improves the performance on benchmark data 
sets. In this section, we survey papers of recent three years on few-shot instance segmentation.
Table. \ref{tab:instance} and Fig. \ref{image} show the research progress of the few-shot instance segmentation.

\begin{table*}
	\caption{The latest performance of FSL in the tasks of few-shot instance segmentation during 2018-2021.}
	\label{tab:instance}
	\resizebox{\linewidth}{!}{
		\begin{tabular}{llccllcc}
			\toprule
			Ref & Model & 1-shot MAP50 & Extra Training Data & Core idea & Approach Taxonomy & Pubilsed Date & Available code\\
			\midrule
			Ganea et al. \cite{ganea2021incremental} & 	
			ONCE iMTFA & 20.13 & \XSolidBrush & \begin{tabular}[c]{@{}l@{}} Learning discriminative \\embedding vectors\end{tabular}  & Metric learning & Jun 2021 & \CheckmarkBold \\
			Nguyen et al. \cite{nguyen2021fapis}& 	
			FAPIS & 16.3 & \XSolidBrush & Modeling of shared parts  & Metric learning & Apr 2021 & \CheckmarkBold \\
			Fan et al. \cite{fan2020fgn} & 	
			FGN  & 16.2 & \XSolidBrush & \begin{tabular}[c]{@{}l@{}} Using attention and\\ relationships to guide\\ generalization\end{tabular} & Based model & Sep 2020 & \CheckmarkBold\\
			He et al. \cite{zhang2020sg} & 	
			Mask R-CNN & 14.8 & \XSolidBrush & \begin{tabular}[c]{@{}l@{}} Adding masks to\\ predicted objects\end{tabular}  & Data Augmentation & Aug 2017 & \CheckmarkBold\\
			Michaelis et al. \cite{michaelis2018one} & 	
			Siamese Mask R-CNN & 14.5 & \XSolidBrush & \begin{tabular}[c]{@{}l@{}} Encoding reference image \\subjects\end{tabular}  & Metric learning  & Nov 2018 & \CheckmarkBold\\
	
			\bottomrule
		\end{tabular}
	}
\end{table*}

\section{Future Direction and Opportunities of FSL }
\label{secop}
Considerable recent work has made promising progress on various task settings for FSL. Nonetheless, for more challenging scenes, both the training and validation data sets are minimal, where the distribution of other data neither helps real samples to be evaluated nor has extensive training data or validation datasets for transfer learning. Moreover, meta-learning also does not have enough tasks to initialize the parameters. With the taxonomy proposed in this survey, in this section we put forward several possible future research directions in FSL. Furthermore, recent advances in applications and algorithms are also presented through this comprehensive survey of FSL.

\subsection{Better evaluation of data distribution}
The essence of FSL is that the support data sets are too small to evaluate the true data distribution. So what exactly can be done to maximize the evaluation of the true data distribution using a limited number of samples? The latest work \cite{yang2021free} is making a useful attempt in this direction, proposing the idea of distribution correction where the mean and covariance of the base class are computationally corrected and then a linear classifier can be used directly to obtain good results.  In fact, the difference between FSL and traditional deep learning is not big enough when the few samples are accurate enough to estimate the true data distribution. This is an exciting direction to explore. Similarly, in the field of computer vision, there are no task settings or datasets based on real application scenarios for FSL. Most of the work is still focused on leveraging and mining information from image data. The current mainstream benchmark datasets have more or less various problems: the mini-Imagenet dataset has some inappropriate samples or too difficult samples, such as solid occlusion, multiple objects in the same image, etc. The Omniglot dataset is far away from practical applications and is not easily inspired in real-world applications. BSCD-FSL \cite{guo2020broader} provides a more violent cross-domain FSL benchmark dataset involving satellite images, medical images. Until now there is no benchmark dataset to evaluate the generalization ability of a model at a fine grained detail. Developing and completing a benchmark dataset in the field of FSL will provide a more realistic evaluation of the current state-of-the-arts in FSL.  

\subsection{Improving the robustness of data-to-label mapping}
 A new challenge to FSL is posed by the emergence of BSCD-FSL. Its emergence explores and reveals the limitations of current FSL solutions for cross-domain learning. Recent research has produced some excellent results in this area, such as skillfully designed task tuning, more sophisticated hyperparameter tuning, formation of auxiliary data sets, and extraction of domain-irrelevant features. Currently, fine-tuning is already performing very robustly at the intersection of transfer learning and meta-learning. Nonetheless, both techniques are still very distinct. Pre-training can be seen as learning many categories of tasks, but it is single-task learning. Meta-learning, on the other hand, is a multi-task learning approach. Whether there is a better model that can integrate meta-learning and fine-tuning to maximize the performance of the model while reducing the computational complexity of meta-learning is a direction worthy of deeper discussion and exchange among researchers at the moment.

\subsection{Learn meta-knowledge more effectively from historical tasks}
Meta learning is still limited to performance in a specific task space under a defined network structure. In the case of classification tasks, only associations between classification tasks can currently be considered. Is it possible to have a framework that can take into account tasks such as classification, detection, prediction, and generation at the same time? This would enable meta-learning to be somewhat separated from the conception of tasks. Some recent work has attempted to optimize each small batch as a whole. In this case, how to optimize the inner loop will be an important direction of optimization for efficient applications. In the future, pre-training and fine-tuning will become the mainstream algorithms for FSL. At present, meta-learning is still exploring the correlation between tasks, and no relevant theory has yet emerged to explain the causal relationship behind meta-learning. As the causation theory framework evolves, meta-learning would probably tend to become a more general framework.

\subsection{Full convergence of multimodal information}

Multimodal learning is currently an emerging approach for solving FSL problems by automatically learning small sample tasks in edge scenarios without supervised information and quickly migrating to data from different domains. It is widely regarded as a path exploration from weak AI in limited domains towards general AI. The implementation of pre-training and fine-tuning in multimodal learning scenarios can largely enable the usage of a uniform feature representation across different tasks. For instance, cross-modal understanding, and cross-modal generation. The emergence of multimodal pre-training models can support multiple tasks, generalize across many scenarios, and have a substantial ability to generalize and replicate at scale. Extensive work has been done on fusing two or more types of information, including semantic information. Nonetheless, the main work is still focused on pixels and semantic information, with a relatively single function. In order to effectively address feature reuse under multiple modalities and reduce the cost of data annotation, there is an urgent need for the industry to materialize a powerful pre-trained model involving the fusion of three and more modalities.

\section{CONCLUSION}
As an important branch of deep learning, few-shot Learning does not require a large amount of data but chooses a softer approach to solve problems, where it can be perfectly integrated with techniques such as transfer learning, meta-learning and data augmentation. In this paper, we provide a comprehensive survey of FSL in the form of questions and answers that easily distinguish the confused concepts and summarize the rich baseline dataset under FSL. Besides, we provide unique insights into the challenges in the development of FSL following a new taxonomy. The evolution of relevant research methods is analyzed in depth according to the degree of integration of knowledge in each stage. Furthermore, for the sake of completeness of the exposition, we also compare and analyze the recent advances of FSL in the field of computer vision. Finally, we present a list of possible future research directions and opportunities in light of the extensive recent literature. Overall, this paper provides an overall comprehensive summary of the frontier advances in FSL over the past three years and is expected to contribute to the synergistic development of FSL and its related fields.

%Few-shot Learning (FSL), as a more challenging machine learning task, can be perfectly integrated with techniques such as transfer learning, meta-learning and data augmentation. As an important branch of deep learning, it does not require a high amount of data but chooses a softer approach to solve problems. We provide a comprehensive survey of FSL, especially in the form of questions and answers that distinguish easily confused concepts and summarize the rich baseline dataset under FSL. We provide insights into the challenges in the development of FSL following a new taxonomy. The evolution of relevant research methods is analyzed in depth according to the degree of integration of knowledge in each stage. In addition, for the sake of completeness of the exposition, we also compare and analyze the recent advances of FSL in the field of computer vision. Finally, we present a list of possible future research directions and opportunities in light of the extensive recent literature. Overall, this paper provides an overall comprehensive summary of the frontier advances in FSL over the past three years and is expected to contribute to the synergistic development of FSL and its related fields.

%\bibliographystyle{ACM-Reference-Format}
%\bibliography{sample-base}

\bibliographystyle{unsrt} 
\bibliography{sample-base}

%\appendix



%Lorem ipsum dolor sit amet, consectetur adipiscing elit. Morbi
%malesuada, quam in pulvinar varius, metus nunc fermentum urna, id
%sollicitudin purus odio sit amet enim. Aliquam ullamcorper eu ipsum
%vel mollis. Curabitur quis dictum nisl. Phasellus vel semper risus, et
%lacinia dolor. Integer ultricies commodo sem nec semper.


%Etiam commodo feugiat nisl pulvinar pellentesque. Etiam auctor sodales
%ligula, non varius nibh pulvinar semper. Suspendisse nec lectus non
%ipsum convallis congue hendrerit vitae sapien. Donec at laoreet
%eros. Vivamus non purus placerat, scelerisque diam eu, cursus
%ante. Etiam aliquam tortor auctor efficitur mattis.


%Nam id fermentum dui. Suspendisse sagittis tortor a nulla mollis, in
%pulvinar ex pretium. Sed interdum orci quis metus euismod, et sagittis
%enim maximus. Vestibulum gravida massa ut felis suscipit
%congue. Quisque mattis elit a risus ultrices commodo venenatis eget
%dui. Etiam sagittis eleifend elementum.

%Nam interdum magna at lectus dignissim, ac dignissim lorem
%rhoncus. Maecenas eu arcu ac neque placerat aliquam. Nunc pulvinar
%massa et mattis lacinia.

\end{document}